\documentclass{article}
\usepackage{iclr2027_conference,times}
\usepackage{amsmath,amssymb,booktabs,multirow,graphicx,microtype,listings}
\usepackage{hyperref}
\usepackage{url}
\usepackage{xcolor}
\usepackage{enumitem}
\usepackage[normalem]{ulem}
\iclrfinalcopy
\title{XOR-Trellis: Ultra-Low-Complexity Dequantization and Curvature-Aware Hadamard-Free LLM Quantization}

\author{
Xiaofan Que$^{1}$, Nir Elkayam$^{1}$, Spandan Pyakurel$^{2}$\thanks{Work done during an internship at Arm.}, Shuokai Pan$^{1}$, Dibakar Gope$^{1}$ \\
$^{1}$Arm, Inc. \hspace{8pt} $^{2}$Rochester Institute of Technology \\
\texttt{\{xiaofan.que, nir.elkayam, shuokai.pan, dibakar.gope\}@arm.com}
\\
\texttt{sp1468@rit.edu}
}

\newcommand{\R}{\mathbb{R}}

\begin{document}
\maketitle

\begin{abstract}
Trellis-coded quantization enables high-dimensional compression of large
language model (LLM) weights at ultra-low bit widths without the exponentially
large codebooks required by conventional vector quantization. Practical
deployment, however, presents two challenges: reconstructing compressed weights
at sufficient parallel throughput to avoid making dequantization an inference
bottleneck, and maintaining quantization accuracy without costly incoherence
transformations. We address these challenges with two complementary techniques.
First, we introduce an ultra-low-complexity trellis dequantizer that uses a
structured, hardware-efficient state-to-value mapping while preserving diverse
reconstruction choices for trellis search. Second, we reformulate discrete
trellis path optimization with a curvature-aware objective that reflects model
sensitivity directly in the original coordinate space. Together, these
techniques enable high-quality ultra-low-bit trellis quantization with
inexpensive, highly parallel runtime reconstruction and without relying on
Hadamard-based incoherence processing.
\end{abstract}

\section{Introduction}

Large language models (LLMs) impose substantial storage and memory-bandwidth
costs during inference, making weight-only post-training quantization (PTQ)
particularly important for autoregressive decoding. At extreme bit widths,
however, maintaining accuracy becomes difficult. Vector quantization (VQ)
improves representational efficiency by jointly quantizing multiple weights,
but conventional VQ codebooks grow exponentially with vector dimension.
Trellis-coded quantization (TCQ) avoids this growth by coupling quantization
decisions through a finite-state path while storing only low-bit transitions,
enabling effective dimensions far beyond conventional VQ
\citep{tseng2024qtip}.

High-dimensional trellis quantization creates an important systems
challenge: compressed representations must be converted back into numerical
weights as they are consumed during inference. Although ultra-low-bit
quantization substantially reduces weight-memory traffic, realizing this
bandwidth advantage requires dequantization to keep pace with the parallel
weight-consumption rate of modern GPUs, NPUs, vector CPUs, and other
accelerators; otherwise, per-weight reconstruction arithmetic can become a
throughput bottleneck and offset part of the memory-bandwidth benefit of
compression. This is particularly challenging for computed trellis
dequantization: for example, QTIP's 3INST dequantizer requires arithmetic,
logical, and data-manipulation operations for each reconstructed weight, which
can require substantial execution bandwidth on programmable processors or
replicated datapath resources in custom accelerators. Our objective is therefore
to make trellis dequantization sufficiently simple and parallelizable to scale
with accelerator weight-consumption bandwidth, without sacrificing the
quantization accuracy enabled by high-dimensional trellis coding. This
hardware--accuracy co-design objective motivates our first contribution.

A second challenge arises from incoherence processing used in a number of
state-of-the-art ultra-low-bit quantization methods. Such methods often apply
randomized Hadamard transforms (RHTs), learned rotations, or related
preprocessing to spread weight outliers and model-sensitive directions more
uniformly, producing representations that are easier for highly constrained
low-bit quantizers to encode. This issue arises not only in trellis-coded
quantization, but more broadly in structured high-dimensional quantization
approaches whose effectiveness can depend on the geometry of the input
representation. QTIP, for example, uses RHT-based incoherence processing to
make weights and important Hessian directions more incoherent and thereby
improve trellis quantization quality \citep{tseng2024qtip}. Rotation-based
methods such as QuaRot and SpinQuant similarly show that rotations can suppress
outliers and improve low-bit quantization, while non-absorbable rotations can
introduce additional inference-time overhead
\citep{ashkboos2024quarot,liu2025spinquant}. Recent LLVQ results show that
structured high-dimensional quantization can reduce dependence on Hadamard
preprocessing \citep{vanderouderaa2026llvq}, motivating the question of whether
trellis quantization can similarly achieve accurate quantization in the
original coordinate space and eliminate the need for Hadamard transformations
in the inference path.

We find that sensitivity to RHT need not be intrinsic to the trellis
representation, but can instead arise in part from a mismatch between
model-aware quantization and discrete reconstruction search. Second-order PTQ
weights errors according to their predicted model impact, whereas conventional
trellis path selection uses isotropic Euclidean distortion. RHT can partly mask
this mismatch by making sensitivities more uniform, so that Euclidean distance
becomes a better surrogate for the model-aware objective. Recent work such as
YAQA similarly highlights the importance of aligning quantization decisions
with appropriate Hessian information \citep{tseng2026yaqa}.

This observation motivates our second contribution: curvature-aware trellis
path optimization in the original, unrotated coordinate space. Rather than transforming the representation to make an isotropic search objective more appropriate, we make the search itself reflect the nonuniform model sensitivity exposed by second-order quantization. We derive an additive curvature-aware trellis path-search objective from the residual sensitivity structure of the LDL factorization, directing reconstruction choices toward errors that are more consequential to the model. The method operates entirely offline and requires no change to the compressed representation or runtime dequantizer.

These observations lead to two complementary contributions:
\begin{itemize}[leftmargin=*]
    \item \textbf{Ultra-low-complexity trellis dequantization.}
    We introduce an XOR-Trellis FP4 dequantizer whose state-to-value mapping is
    jointly designed for quantization quality and hardware efficiency. Reconstruction
    uses shallow parity/XOR operations and a small fixed FP4 palette lookup rather
    than a large state-to-value table or expensive computed function. The resulting structure scales to wide parallel dequantization so that
    dequantization throughput can track the weight-consumption bandwidth of the
    target processor, while structured local palettes preserve the reconstruction
    diversity required for accurate trellis search.
    \item \textbf{Curvature-aware rotation-free trellis optimization.}
    We identify a fundamental mismatch between the model-aware objective of
    second-order quantization and the isotropic distortion used for discrete trellis
    path selection. We introduce a curvature-aware Viterbi objective that uses the
    residual sensitivity structure exposed by the LDL factorization, directly
    accounting for nonuniform sensitivity in the original coordinate space. This
    enables accurate rotation-free trellis quantization without changing the
    compressed representation or increasing runtime dequantization complexity.
    \end{itemize}
    
On Llama-3.1-8B, the resulting curvature-aware XOR-Trellis method achieves a
WikiText-2 perplexity of $8.00$ directly in the original coordinate space,
substantially improving over conventional unrotated trellis quantization
($8.69$) and outperforming the RHT-enabled QTIP baseline ($8.48$).


\section{Background and Motivation}
\subsection{From vector quantization to trellis quantization}
Scalar quantization independently maps each weight to a reconstruction level,
whereas VQ maps a vector $\mathbf{w}\in\R^d$ to a codeword $\mathbf{c}$,
improving shaping efficiency by balancing errors jointly across dimensions.
However, maintaining a fixed bitrate as $d$ increases requires exponentially
more codewords.
Trellis-coded quantization (TCQ) instead represents a sequence of quantization
decisions as a path through a finite-state trellis. At each step, the current
state and a low-bit branch determine both a reconstruction value and the
subsequent state. Quantization therefore searches candidate paths, typically
using the Viterbi algorithm, to find a sequence of reconstruction values that
minimizes a distortion objective. Coupling consecutive reconstruction decisions
through the trellis state provides high-dimensional shaping without explicitly
storing a high-dimensional VQ codebook.

QTIP applies TCQ to ultra-low-bit LLM weight quantization using a bitshift
trellis and state-dependent reconstruction mapping \citep{tseng2024qtip}. We
denote by $K$ the number of transition bits per trellis step, giving $2^K$
possible branches. For the $K=2$ trellis considered here, each step therefore
selects one of four two-bit branches. A finite state retains recent branch
history, so the reconstruction choices for the current weight depend on earlier
decisions. Our example uses a 16-bit state comprising 14 retained
history bits and the current two-bit branch. Each selected branch enters the
history as the oldest bits are discarded, causing the state to act as a sliding
memory of recent quantization decisions.
\subsection{LDLQ and Hessian information}
\label{sec:ldlq-background}
Second-order PTQ methods use an approximate Hessian $H$ to describe how
sensitive the model is to quantization errors
\citep{frantar2022gptq,chee2024quip}. LDLQ-style quantization factorizes
\begin{equation}
    H = L D L^\top,
    \label{eq:ldl}
\end{equation}
which separates this sensitivity information into two useful components.
The triangular factor $L$ captures how quantization errors in different weight
dimensions interact and is already used by LDLQ for sequential error feedback
and compensation. The diagonal factor $D$ describes the relative importance
of the remaining errors after these interactions have been accounted for.
Thus, a large $D_{ii}$ indicates that an error in dimension $i$ is more costly
to the model than an equal-magnitude error in a dimension with a smaller
$D_{ii}$.

\subsection{Why rotations help---and why we want to remove them}
RHT-based incoherence processing mixes coordinates so that weight outliers and
sensitivity are spread more uniformly, improving the fit of constrained
low-bit quantizers \citep{tseng2024qtip}. QuaRot and SpinQuant similarly
demonstrate the value of rotations for suppressing outliers in LLM quantization
\citep{ashkboos2024quarot,liu2025spinquant}. Although weight rotations can be
prepared offline, rotations that cannot be absorbed into surrounding operations
require runtime transformation and introduce additional inference-time overhead.
As motivated above, RHT can also make Euclidean trellis path distortion a
better surrogate for model-aware quantization by reducing sensitivity
nonuniformity. Section~\ref{sec:hessian-search} develops our alternative:
making the path-search objective explicitly sensitivity-aware rather than
transforming the representation, thereby avoiding online RHT processing and
its associated runtime overhead, which can otherwise limit LLM inference
throughput. Further related work is discussed in
Appendix~\ref{app:related_works}.

\section{XOR-Trellis FP4 Trellis Decoder}
\label{sec:xor-decoder}
\subsection{Design objective: accurate trellis search with inexpensive reconstruction}
Trellis-coded quantization couples reconstruction decisions across a sequence through a finite-state path. Unlike scalar quantization, the reconstruction choices available for the current weight depend on preceding quantization decisions. This path dependence provides some of the high-dimensional shaping benefit of vector quantization without explicitly materializing an exponentially large high-dimensional codebook.

The same property creates a runtime reconstruction problem. Each compressed transition must be combined with the current trellis state to determine the corresponding numerical weight. A large state-to-value lookup is unattractive, while computed-code approaches replace storage with arithmetic that is exercised repeatedly over large LLM weight tensors. We therefore seek a dequantizer (generator) whose state-to-value mapping is simultaneously \emph{expressive enough to preserve accurate trellis search} and \emph{simple and parallelizable enough for high-throughput runtime reconstruction}.

\subsection{Trellis state and local reconstruction palettes}
For the $K=2$ bitshift trellis considered here, two bits encode one of four branches at each step. The trellis state acts as a finite memory of recent branch decisions: after each weight is quantized, its two-bit transition is incorporated into the sliding state while the oldest transition bits are discarded. In the demonstrated configuration, the state is 16 bits, comprising a retained 14-bit history $h$ and a current two-bit branch $b\in\{0,1,2,3\}$. This history couples neighboring quantization decisions and determines which reconstruction choices are locally available.

For a fixed history $h$, the four branches define a \emph{local reconstruction palette}: the four values Viterbi can choose at that search step. This conditional four-way set is more informative than the generator's global output histogram alone. For example, a local palette $[-2,-2,2,2]$ nominally exposes four branches but only two distinct reconstructions, wasting half of the local representational freedom. We therefore design the generator around the geometry of these conditional palettes. In this sense, the generator behaves like a tiny state-dependent codebook whose contents change with trellis history.

\subsection{XOR-Trellis mapping}


In the FP4 embodiment evaluated in this work, we use four
reconstruction palettes:
\begin{align*}
P_0 &= [-6,-1.5,0,2], & P_1 &= [-4,-1,0.5,3],\\
P_2 &= [-3,-0.5,1,4], & P_3 &= [-2,0,1.5,6].
\end{align*}
Every palette contains four numerically distinct values spanning negative,
near-zero, and positive regions, with minimum spacing 1.5. Consequently,
each of the $2^{14}=16{,}384$ retained histories exposes four distinct,
well-separated reconstruction choices.

These palettes provide one concrete realization of the proposed
state-dependent reconstruction mapping; the dequantizer itself is not
restricted to FP4 or to four-bit reconstruction values and can be
instantiated with other numerical formats and corresponding reconstruction
palettes\footnote{Although we report results only for the FP4 (E2M1) format, preliminary
investigations with FP8 (E4M3) using 16 reconstruction palettes and correspondingly
optimized palette values suggest slightly better reconstruction error than the FP4
variant, providing evidence that XOR-Trellis generalizes beyond FP4.}.

The palette construction is not merely heuristic. Under a fixed inventory
consisting of the 16 labeled FP4 payloads, uniform use of four disjoint
four-entry palettes, and scalar squared-error distortion, the rank-interleaved
palettes above are pointwise optimal for the immediate four-way reconstruction
choice: for every scalar target $x$, they attain the minimum possible average
distance to the nearest entry across the four palettes. They also achieve the
maximum possible minimum within-palette numerical separation of 1.5.
Appendix~\ref{app:palette-optimality} states these results formally and
provides the proofs.

Four masked parity functions of the retained history produce bits
$p_0,p_1,q_0,q_1$, where each function XOR-reduces a subset of history
bits specified by a corresponding binary mask. The first pair determines
which palette is active, while the second pair permutes how the four branch
identities map to palette entries:
\begin{equation}
p=p_0|(p_1\ll1),\qquad
b'=b\oplus(q_0|(q_1\ll1)),\qquad
\hat w=P_p[b'].
\label{eq:xor-decoder}
\end{equation}
We use masks $\{\texttt{0x1089},\texttt{0x2112},\texttt{0x0A24},\texttt{0x2441}\}$ to compute $p_0,p_1,q_0,q_1$, respectively; other mask choices can realize the same general XOR-Trellis construction. Palette selection controls the local reconstruction geometry, while branch permutation varies how branch identities are associated with reconstruction values. Appendix~\ref{app:finite-state} formalizes structural properties of this XOR mapping over $\mathbb F_2$. In particular, binary-rank conditions characterize palette marginal balance,
and the presence or absence of deterministic constraints in multi-step palette sequences. These conditions provide exact tests for structural degeneracies in candidate XOR mappings while leaving final mask selection as a finite-horizon coding problem.
\begin{figure*}[t]
    \centering
    \includegraphics[width=0.9\textwidth]{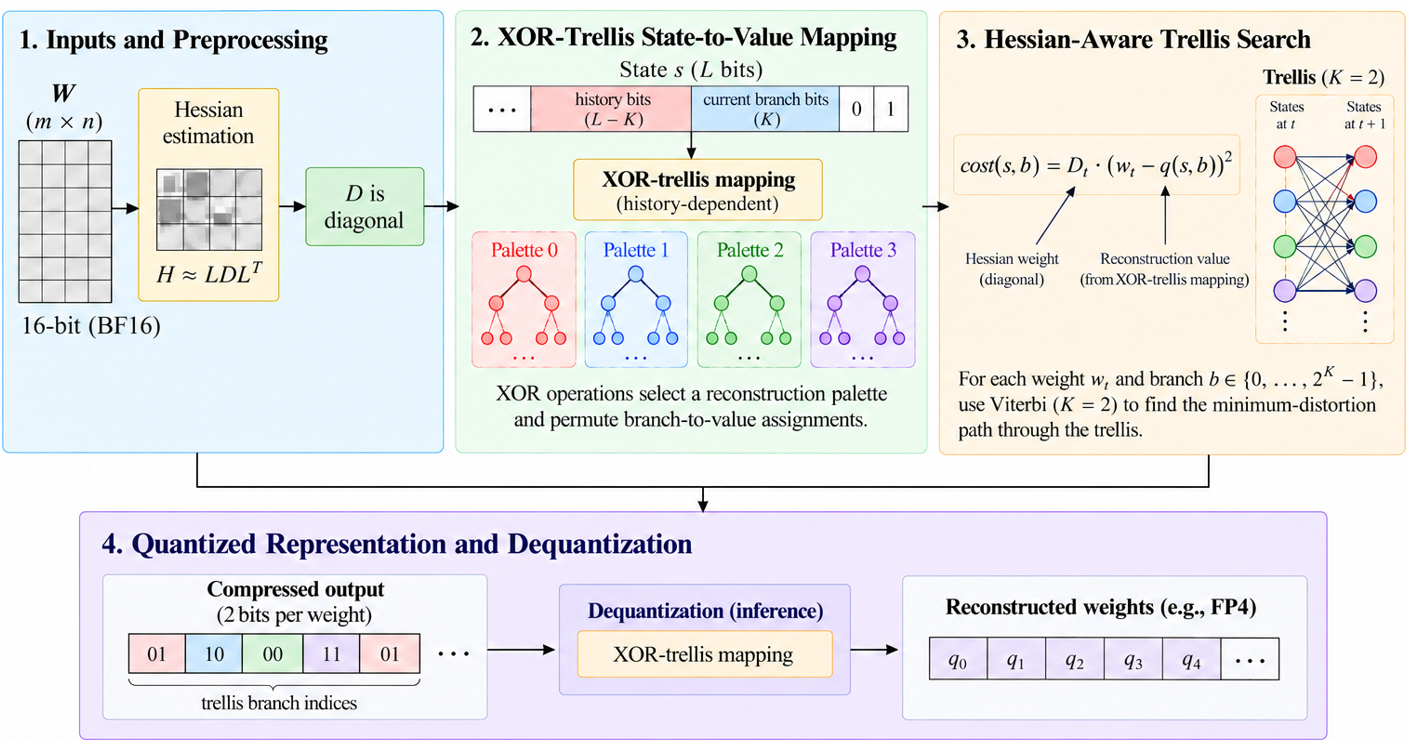}\vspace{-2mm}
    \caption{
Overview of XOR-Trellis. A structured state-to-value mapping provides diverse FP4 reconstruction choices, while Hessian-aware Viterbi search weights errors using the diagonal factor $D$. The resulting approximately 2-bit trellis path is decoded with the same lightweight mapping.
}
    \label{fig:xor_stratified_methodology}
\end{figure*}\vspace{-2mm}
\subsection{Local geometry and future search freedom}
Local diversity matters beyond the immediate reconstruction error because the selected branch becomes part of the history used at subsequent steps. A locally poor branch set can therefore affect not only the current approximation but also the reconstruction choices reachable later in the trellis. The XOR-Trellis construction combines four distinct local values with history-dependent branch permutation, avoiding wasted local branches while varying how reconstruction values are associated with subsequent state evolution. We refer to this design objective as preserving \emph{future search freedom}: the trellis should avoid wasting nominal branches on numerically equivalent local choices or imposing avoidable deterministic restrictions on future reconstruction sequences. Appendix~\ref{app:xor-rank-theory} formalizes this intuition through multi-step rank conditions that detect restricted palette-transition and longer-window patterns.

The global reconstruction distribution remains important, but it is complementary to this local criterion. Exhaustively decoding all $2^{16}=65{,}536$ states spans the available FP4 reconstruction range and produces a broad distribution with substantial tail mass. A favorable global histogram alone, however, does not guarantee that the four choices presented to Viterbi at a particular history are useful; both global coverage and conditional local geometry matter.

\subsection{State utilization and path mixing}
We next examine how effectively the XOR-Trellis mapping utilizes the information retained in the trellis state. In exhaustive analysis of the demonstrated XOR-Trellis construction, flipping any evaluated state bit changes the decoded numerical value, indicating that all state bits participate meaningfully in the demonstrated mapping. In addition, signed and absolute-value linear correlations between a reconstruction and the reconstructions reachable by successor transitions are essentially zero. These empirical diagnostics complement the algebraic analysis in Appendix~\ref{app:palette-optimality}, where rank conditions characterize marginal, adjacent-transition, and multi-step balance of the XOR mapping. Decorrelation alone is not sufficient for good quantization, nor do the rank conditions establish global optimality; together with distinct local palettes, however, they provide complementary evidence that the trellis state is used to create diverse reconstruction trajectories.

\subsection{Hardware realization}
A 16-bit state nominally defines $2^{16}=65{,}536$ possible state-dependent
reconstruction outcomes, but Equation~\ref{eq:xor-decoder} does not require a
correspondingly large state-to-value lookup. Instead, the mapping factorizes
into four shallow masked-parity reductions, two XOR operations for branch
permutation, and a small fixed $4\times4$ FP4 palette lookup. In the evaluated
gate-level realization, the pre-lookup logic requires approximately 14 XOR2
gates. Thus, high-dimensional trellis coupling can be paired with an extremely
small per-weight dequantization datapath.
This property is important for deployment. The shallow XOR-dominated structure is amenable to wide parallel implementation, allowing many compressed transitions to be reconstructed concurrently into FP4 weights immediately before or adjacent to the arithmetic datapath. This allows dequantization throughput to scale with the parallel weight-consumption rate of modern neural-network accelerators while retaining a compact, area- and energy-efficient reconstruction datapath.

\section{Curvature-Aware Trellis Path Optimization Without RHT}
\label{sec:hessian-search}
\subsection{Second-order PTQ and trellis search optimize different geometries}
Trellis-coded PTQ combines two optimization procedures whose objectives are not,
in general, aligned. The outer second-order quantization procedure seeks to
minimize a model-aware quadratic proxy for the effect of weight perturbations.
For quantization error $e=w-\hat w$, a common local objective has the form
\begin{equation}
    \mathcal{L}_{Q}(e) \propto e^\top H e,
    \label{eq:hessian-proxy}
\end{equation}
where $H$ is an approximate Hessian derived from layer activations. Equal Euclidean errors can therefore have very different model impact depending on direction.

The inner trellis optimization is discrete. Given LDLQ-adjusted targets $z_i$ and path-dependent reconstructions $\hat z_i(\pi)$, conventional Viterbi search selects
\begin{equation}
    \pi_{\mathrm{Euc}}
    \in \arg\min_{\pi}\sum_i\bigl(z_i-\hat z_i(\pi)\bigr)^2.
    \label{eq:euclidean}
\end{equation}
Thus, the sequential second-order optimizer and the discrete path search use
different error metrics: the former accounts for model sensitivity, whereas the
latter treats residual errors isotropically. We refer to this as the
\emph{second-order/path-search objective mismatch}. A path that is optimal in Euclidean reconstruction error need not be the path that best preserves the model under the local quadratic metric.
Our central hypothesis is that incoherence processing partly masks this mismatch. By mixing coordinates, RHT makes weights and sensitivity more homogeneous, so an isotropic path metric becomes a better surrogate. Once RHT is removed, the mismatch becomes substantially more important.

\subsection{LDL factorization exposes residual curvature}
Using the LDL factorization introduced in Section~\ref{sec:ldlq-background}, the second-order objective
can be rewritten as
\begin{equation}
    e^\top H e=(L^\top e)^\top D(L^\top e).
    \label{eq:ldl-residual-main}
\end{equation}
Thus, after the error interactions represented by $L$ are accounted for, $D$
provides a diagonal measure of the relative cost of the remaining error
directions. This residual sensitivity information motivates the
curvature-aware trellis objective developed below.
Conventional trellis path selection does not explicitly preserve this second role: after LDLQ constructs feedback-adjusted targets, Euclidean Viterbi effectively uses an identity metric over the residuals presented to the discrete search. Appendix~\ref{app:hessian-viterbi-theory} formalizes this interpretation, including the exact LDL quadratic identity, an anisotropy-dependent approximation bound, and the weighted Viterbi recurrence.

\subsection{Curvature-aware Viterbi objective}
The factorization suggests a curvature-aware objective for discrete trellis path selection. 
We retain the LDLQ-adjusted targets, trellis topology, state-to-reconstruction
mapping, and dynamic program, but optimize
\begin{equation}
    \pi_D
    \in \arg\min_{\pi}
    \sum_i D_{ii}\bigl(z_i-\hat z_i(\pi)\bigr)^2.
    \label{eq:diagD}
\end{equation}
The objective restores coordinate-dependent diagonal curvature to path selection while remaining additive over trellis transitions. Equivalently, each residual is scaled by $\sqrt{D_{ii}}$ before squared distortion is accumulated.
The effect is intuitive. Suppose two candidate residuals have equal magnitude, $|e_i|=|e_j|=0.1$, but $D_{ii}=8D_{jj}$. Euclidean path search assigns the same cost to both; the curvature-aware objective assigns the first residual eight times the cost and therefore encourages the trellis to spend its limited discrete reconstruction choices protecting the more sensitive direction.

Importantly, the simplicity of the implementation is a consequence of the factorization rather than the motivation for the method. Realizing the curvature-aware objective requires only a scalar weighting of each existing branch distortion. The state space, branch structure, survivor count, compressed representation, and asymptotic Viterbi complexity are unchanged. Unlike YAQA, which improves adaptive rounding by constructing Hessian approximations that better target end-to-end model error~\citep{tseng2026yaqa}, we retain the existing Hessian approximation and LDLQ feedback rule and make the discrete trellis optimization itself curvature-aware.

\subsection{RHT as conditioning; curvature awareness as an alternative}
RHT and curvature-aware path search address a common difficulty from different directions:
\begin{equation*}
\begin{array}{lll}
\text{RHT} &:& \text{change the representation so an isotropic path metric is more appropriate},\\[2pt]
\text{Ours} &:& \text{retain the representation and make the path metric reflect its curvature}.
\end{array}
\end{equation*}
RHT mixes the coordinate system before quantization, spreading large weights and sensitive directions and making the transformed problem more homogeneous. In that regime, an unweighted Euclidean metric can be an effective approximation. 
Without RHT, residual curvature can be strongly anisotropic, and equal weighting
can mis-rank candidate trellis paths relative to the curvature-aware objective.

This interpretation yields a concrete prediction. If $D=\alpha I$ for some $\alpha>0$, then curvature-aware and Euclidean Viterbi have exactly the same minimizers because their objectives differ only by a positive constant. More generally, Appendix~\ref{app:hessian-viterbi-theory} shows that the quality of the Euclidean surrogate under the diagonal-curvature objective is controlled by the diagonal anisotropy $\kappa_D=D_{\max}/D_{\min}$. The approximation is tightest near isotropy and can weaken as the spread of $D$ grows.

This also explains why we do not treat $D$-weighting as a universal replacement for the original metric. Our experiments favor
\begin{equation}
    d_i =
    \begin{cases}
        D_{ii}(z_i-\hat z_i)^2, & \text{without RHT},\\[2pt]
        (z_i-\hat z_i)^2, & \text{with RHT},
    \end{cases}
    \label{eq:policy}
\end{equation}
because the unweighted metric remains preferable after RHT has already conditioned the problem.

\subsection{Mechanistic evidence: anisotropy predicts where metric awareness matters}
We test the above prediction across all 224 Q/K/V/O/up/gate/down projection matrices of Llama-3.1-8B. Using the robust anisotropy statistic $A_D=P_{95}(D_{ii})/P_5(D_{ii})$, we find that anisotropy is positively associated both with the benefit of RHT under conventional Euclidean trellis search and with the gain from curvature-aware search when RHT is removed. In particular, the Spearman correlation between $\log A_D$ and the no-RHT diagonal-$D$ gain is $\rho=0.545$, while its correlation with the legacy RHT gain is $\rho=0.629$. The corresponding absolute legacy error reduction has $\rho=0.606$.

The common dependence supports the proposed mechanism: RHT is most useful to
conventional Euclidean path search on matrices where the residual metric is
highly nonuniform, and explicit curvature weighting is also most useful there
when RHT is absent. The relationship is not complete, however. Diagonal anisotropy alone does not determine whether no-RHT $D$-weighting beats the full RHT pipeline for every matrix, because RHT can also redistribute off-diagonal coupling and interact with projection-specific structure. Appendix~\ref{app:d-anisotropy} gives the full protocol, confidence intervals, counterexamples, and matrix-level analysis.\vspace{-1mm}

\subsection{Further Exploiting Curvature through Energy Ordering}\vspace{-2mm}

Beyond weighting the path-search objective, the same curvature information can
also guide the order in which LDLQ processes feature blocks. We reorder feature
blocks according to their aggregate curvature before quantization. For each block $\mathcal{B}_b$, we define its energy as
\[
E_b = \operatorname{tr}\!\left(H_{\mathcal{B}_b,\mathcal{B}_b}\right)
    = \sum_{i \in \mathcal{B}_b} H_{ii},
\]
and process the blocks in descending order of $E_b$. 
This places the most sensitive feature directions early in the LDLQ recursion,
before errors from less important blocks have accumulated. This ordering gives
subsequent blocks the opportunity to compensate for errors introduced while
quantizing higher-curvature blocks through the triangular feedback represented
by $L$.
Energy ordering is complementary to curvature-aware Viterbi optimization:
the latter uses curvature to improve candidate-path scoring within each block,
whereas the former uses curvature to determine the order in which blockwise
errors enter the sequential compensation process.
We consistently permute the weights and Hessian before factorization and restore the original feature order after quantization. As with curvature-aware branch weighting, energy ordering is entirely offline and requires neither additional model bits nor modifications to the runtime decoder.\vspace{-2mm}
\vspace{-2mm}\begin{table}[t]
\centering
\vspace{-8mm}
\caption{ Perplexity (PPL) is evaluated on WikiText-2 (WT2) and C4 with context lengths of 4096 for Llama-3.1-8B, 2048 for Llama-3.1-8B-Instruct, and 8192 for Llama-3.1-70B-Instruct. Zero-shot raw accuracy (\texttt{acc}, \%) is reported on five downstream tasks.  }
\label{tab:llama31_all_results}
\resizebox{\columnwidth}{!}{%
\begin{tabular}{lllcc|cccccc}
\toprule
& & & \multicolumn{2}{c|}{PPL $\downarrow$}
& \multicolumn{6}{c}{0-Shot Raw Accuracy (\%) $\uparrow$} \\
\cmidrule(lr){4-5}
\cmidrule(lr){6-11}
Model & Method & Rotation & WT2 & C4 & ArcC & ArcE & BoolQ & HSwag & PiQA & Avg. \\
\midrule

\multirow{6}{*}{Llama-3.1-8B}
& BF16
& --
& 5.84 & 8.43
& 51.28 & 81.61 & 82.05 & 60.06 & 80.14 & 71.03 \\

& QTIP-3INST
& Hadamard
& 8.48 & 11.93
& 39.85 & 73.15 & 78.47 & 51.99 & 75.63 & 63.82 \\

& XOR-Trellis
& Hadamard
& 8.10 & 11.45
& 39.80 & 73.15 & 75.26 & 53.61 & 76.88 & 63.74 \\

& XOR-Trellis
& None
& 8.69 & 12.23
& 37.97 & 68.69 & 76.06 & 51.13 & 75.24 & 61.82 \\

& XOR-Trellis + D-weighted
& None
& 8.06 & 11.40
& 38.48 & 72.39 & \textbf{79.63} & \textbf{53.30} & \textbf{76.82} & 64.12 \\

& XOR-Trellis + D-weighted + energy ordering
& None
& \textbf{8.00} & \textbf{11.34}
& \textbf{42.24} & \textbf{75.04} & 78.26 & 53.22 & 75.73 & \textbf{64.90} \\

\midrule

\multirow{6}{*}{Llama-3.1-8B-Instruct}
& BF16
& --
& 7.22 & 10.44
& 51.71 & 81.82 & 84.10 & 59.05 & 79.92 & 71.32 \\

& QTIP-3INST
& Hadamard
& 10.57 & 14.92
& 43.43 & 76.60 & 82.17 & 51.51 & 77.15 & 66.17 \\

& XOR-Trellis
& Hadamard
& 10.06 & 14.31
& 43.00 & 76.94 & 82.45 & 52.07 & 75.90 & 66.07 \\

& XOR-Trellis
& None
& 10.53 & 14.74
& 37.20 & 70.75 & 80.83 & 51.89 & 74.10 & 62.95 \\

& XOR-Trellis + D-weighted
& None
& \textbf{9.64} & \textbf{13.66}
& 43.43 & 76.81 & 79.24 & \textbf{53.26} & \textbf{77.20} & 65.99 \\

& XOR-Trellis + D-weighted + energy ordering
& None
& 9.73 & 13.70
& \textbf{44.45} & \textbf{77.02} & \textbf{82.78} & 53.23 & 76.33 & \textbf{66.76} \\

\midrule

\multirow{6}{*}{Llama-3.1-70B-Instruct}
& BF16
& --
& 3.52 & 6.46
& 63.31 & 83.75 & 87.68 & 84.63 & 83.79 & 80.63 \\

& QTIP-3INST
& Hadamard
& 6.31 & 8.02
& 58.19 & 82.41 & 87.46 & 80.15 & 81.61 & 77.96 \\

& XOR-Trellis
& Hadamard
& 5.80 & 7.69
& 59.47 & 81.44 & 86.91 & 81.48 & 81.39 & 78.14 \\

& XOR-Trellis
& None
& 6.11 & 7.90
& 58.45 & 80.77 & 85.50 & 80.83 & 81.28 & 77.37 \\

& XOR-Trellis + D-weighted
& None
& 5.62 & 7.58
& \textbf{60.84} & \textbf{81.94} & 85.93 & 81.18 & \textbf{82.26} & \textbf{78.43} \\

& XOR-Trellis + D-weighted + energy ordering
& None
& \textbf{5.52} & \textbf{7.41}
& 58.70 & 81.73 & \textbf{86.67} & \textbf{81.55} & 81.07 & 77.94 \\

\bottomrule
\end{tabular}%
}
\vspace{-4mm}
\end{table}
\section{Experiments}\vspace{-2mm}

We evaluate weight-only trellis quantization using the XOR-Trellis FP4 decoder at approximately 2.25 effective bits per weight, including one E5M3 scale per group of 32 weights\footnote{We focus on group-wise FP4 for its native support and high
low-precision matrix-multiplication throughput on current and next-generation
accelerators, including NVIDIA Blackwell \citep{nvidia_nvfp4_blackwell}.
NVIDIA Rubin further supports block-scaled FP4 GEMMs with UE5M3 scales
\citep{nvidia_cutlass_rubin}; see Appendix~\ref{app:related_works}.}. Unless otherwise specified, experiments use $L/K/V=16/2/1$, trellis tiles $T_x=T_y=16$, 384 RedPajama calibration sequences, seed 0, and no layer-wise or end-to-end fine-tuning. Context length is model-specific and reported with each result. We compare RHT and no-RHT configurations, including Hessian-aware Viterbi weighting and energy-ordered LDLQ variants where indicated. We evaluated both perplexity and zero-shot accuracy on quantized Llama, Mistral and Qwen models. We compare our method against the state-of-the-art low-bit quantization methods QTIP~\cite{tseng2024qtip} and YAQA~\cite{tseng2026yaqa}, as well as the state-of-the-art rotation-free quantization method LLVQ~\cite{vanderouderaa2026llvq}.
For QTIP and YAQA, we reproduce the results using their publicly available code and Hessians under the same non-finetuning setting as our method. 

\subsection{Comparison with QTIP}\vspace{-2mm}
In this section, we compare our rotation-free method with QTIP across multiple model families and scales. We present the Llama-3.1 results in the main text in Table~\ref{tab:llama31_all_results}, covering Llama-3.1-8B, Llama-3.1-8B-Instruct, and Llama-3.1-70B-Instruct. Across all three models, directly removing the Hadamard transform degrades XOR-Trellis; for example, on Llama-3.1-8B, WT2 perplexity increases from 8.10 with RHT to 8.69 without RHT, indicating that groupwise scaling alone does not fully compensate for the loss of rotation. Incorporating diagonal-$D$ weighting substantially recovers this degradation while remaining fully rotation-free: WT2/C4 perplexity improves from 8.69/12.23 to 8.06/11.40, and further to 8.00/11.34 with energy ordering, outperforming QTIP-3INST with Hadamard rotation at 8.48/11.93. A similar trend holds for Llama-3.1-70B-Instruct, where diagonal-$D$ weighting reduces perplexity from 6.11/7.90 to 5.62/7.58 and improves average zero-shot accuracy from 77.37\% to 78.43\%. On Llama-3.1-8B-Instruct, it likewise improves WT2/C4 perplexity from 10.53/14.74 to 9.64/13.66. Overall, these results show that curvature-aware trellis search can recover much of the performance lost when RHT is removed, with energy ordering providing additional gains in several settings. Results on additional model families and scales, including Llama-2, Llama-3.2, Qwen3, and Mistral, are deferred to Appendix~\ref{app:add_result}, where we observe the same overall trend. We further investigate the limitations of groupwise scaling without RHT and its relationship to Hessian anisotropy in Appendix~\ref{app:groupwise_scaling}.

\vspace{-2mm}\subsection{Comparison with YAQA}\vspace{-2mm}
Table~\ref{tab:llama31_8b_instruct_yaqa_results} compares our method with YAQA on Llama-3.1-8B-Instruct using the same two-sided YAQA Hessian and no finetuning. YAQA-3INST performs well with RHT but collapses when the rotation is removed, and adding D-weighted alone does not fully recover it. In contrast, XOR-Trellis remains stable without RHT, and combining it with D-weighted achieves 8.96/12.90 PPL and 67.98\% average accuracy, outperforming YAQA-3INST with RHT. These results suggest that the trellis mapping and Hessian-aware weighting play complementary roles in enabling robust rotation-free quantization.\vspace{-1.35mm}
\begin{table}[t]
\centering
\vspace{-8mm}
\caption{Weight-only PTQ results on Llama-3.1-8B-Instruct using the two-sided YAQA Hessian. Perplexity (PPL) is evaluated on WikiText-2 (WT2) and C4 with a context length of 2048, and zero-shot raw accuracy (\texttt{acc}, \%) is reported on five downstream tasks.}
\label{tab:llama31_8b_instruct_yaqa_results}
\resizebox{\columnwidth}{!}{%
\begin{tabular}{lllcc|cccccc}
\toprule
& & & \multicolumn{2}{c|}{PPL $\downarrow$}
& \multicolumn{6}{c}{0-Shot Accuracy (\%) $\uparrow$} \\
\cmidrule(lr){4-5}
\cmidrule(lr){6-11}
Method & Rotation & Hessian & WT2 & C4 & ArcC & ArcE & BoolQ & HSwag & PiQA & Avg. \\
\midrule
BF16
    & -- & --
    & 7.22 & 10.44
    & 51.71 & 81.82 & 84.10 & 59.05 & 79.92 & 71.32 \\

YAQA-3INST
    & Hadamard & YAQA-2S
    & 9.57 & 13.75
    & 44.71 & 77.57 & 83.46 & 52.35 & 76.17 & 66.85 \\

YAQA-3INST
    & None & YAQA-2S
    & 15511.81 & 8986.10
    & 21.93 & 25.59 & 37.83 & 25.81 & 53.48 & 32.93 \\

YAQA-3INST + D-weighted
    & None & YAQA-2S
    & 6500.82 & 3745.74
    & 20.90 & 26.47 & 37.83 & 26.14 & 52.39 & 32.75 \\

\midrule
XOR-Trellis
    & Hadamard & YAQA-2S
    & 9.00 & 12.90
    & 46.16 & 78.70 & 83.82 & 53.85 & 77.20 & 67.95 \\

\midrule
XOR-Trellis
    & None & YAQA-2S
    & 9.61 & 13.63
    & 44.97 & 75.67 & 82.17 & 53.38 & 76.22 & 66.48 \\

XOR-Trellis + D-weighted
    & None & YAQA-2S
    & \textbf{8.96} & 12.90
    & \textbf{47.53} & \textbf{79.08} & 82.17 & \textbf{54.20} & 76.93 & \textbf{67.98} \\

XOR-Trellis + D-weighted + energy ordering
    & None & YAQA-2S
    & 8.99 & \textbf{12.87}
    & 45.31 & 78.41 & \textbf{82.32} & 53.76 & \textbf{77.48} & 67.46 \\
\bottomrule
\end{tabular}%
}
\end{table}

\subsection{Comparison with LLVQ}
Table~\ref{tab:llvq_comparison} compares XOR-Trellis with LLVQ on Llama-2-7B. LLVQ benefits from input/output rotation, while its fully rotation-free configuration is weaker. In our pipeline, removing RHT increases WT2 perplexity from 6.65 to 7.34, but adding diagonal-$D$ weighting reduces it to 6.41 and improves average zero-shot accuracy from 49.34\% to 61.75\%. This rotation-free result outperforms QTIP-3INST with Hadamard rotation and the best reported LLVQ configuration in the table on WT2 perplexity, while also exceeding the LLVQ input+output rotation configuration in average downstream accuracy. Energy ordering gives a comparable 6.45 PPL and 60.23\% average accuracy, further supporting the effectiveness of Hessian-aware rotation-free trellis search.

\begin{table}[t]
\centering
\vspace{-3mm}
\caption{
Comparison with LLVQ on Llama-2-7B. Perplexity (PPL) is evaluated on WikiText-2 (WT2) with a context length of 4096, and zero-shot raw accuracy (\texttt{acc}, \%) is reported where available. All quantized results are reported without layerwise or end-to-end finetuning. }
\label{tab:llvq_comparison}
\resizebox{0.85\columnwidth}{!}{%
\begin{tabular}{llc|cccccc}
\toprule
Method & Rotation & WT2 $\downarrow$ & ArcC & ArcE & BoolQ & HSwag & PiQA & Avg. \\
\midrule
BF16
    & --
    & 5.12
    & 43.43 & 76.35 & 77.74 & 57.11 & 78.07 & 66.54 \\

QTIP-3INST
    & Hadamard
    & 6.95
    & 34.56 & 69.53 & 71.93 & 48.15 & 74.48 & 59.73 \\

\midrule

LLVQ (shape-gain, 2-bit gain)
    & None
    & 7.27
    & -- & -- & -- & -- & -- & -- \\

LLVQ (shape-gain, 2-bit gain)
    & Input
    & 6.90
    & -- & -- & -- & -- & -- & -- \\

LLVQ (shape-gain, 2-bit gain)
    & Input + Output
    & 6.83
    & 35.50 & 69.80 & 73.00 & 49.70 & 75.20 & 60.64 \\

LLVQ (shape-gain, 0-bit gain)
    & Input + Output
    & \underline{6.48}
    & -- & -- & -- & -- & -- & -- \\

\midrule

XOR-Trellis
    & None
    & 7.34
    & 25.17 & 53.87 & 68.41 & 35.87 & 63.38 & 49.34 \\

XOR-Trellis + D-weighted
    & None
    & \textbf{6.41}
    & \textbf{37.12} & \textbf{71.17} & \textbf{75.05} & \textbf{51.16} & 74.27 & \textbf{61.75} \\

XOR-Trellis + D-weighted + energy ordering
    & None
    & 6.45
    & 35.92 & 69.61 & 69.88 & 50.82 & \textbf{74.92} & 60.23 \\

\bottomrule
\end{tabular}%
}
\vspace{-4mm}
\end{table}\vspace{-4mm}

\subsection{Inference complexity}\vspace{-1.35mm}
Ultra-low-bit weight compression provides its full bandwidth benefit only when
dequantization can keep pace with the parallel weight-consumption rate of the
compute engine. The arithmetic reconstruction used by QTIP-3INST is estimated
to require the equivalent of approximately twelve 32-bit add/subtract
operations, an additional 32-bit addition, mask/XOR logic, and an FP16
addition per reconstruction, corresponding to roughly $3$--$5$k small
logic-gate equivalents depending on the FP16-adder implementation.
XOR-Trellis instead reduces the state-to-value mapping to approximately
14 XOR2 (2-bit XOR) gates and a tiny fixed $4\times4$ palette lookup, while achieving
better perplexity in our matched evaluation. Moreover, a 512-weight-per-cycle
implementation of the XOR-Trellis dequantizer occupies only
$1{,}917~\mu\mathrm{m}^2$ when synthesized using N3P standard cells,
demonstrating that the reconstruction function can be replicated at the wide
parallelism required to keep pace with high-throughput weight consumption.

\vspace{-3mm}\section{Conclusion}\vspace{-3mm}
We presented XOR-Trellis, a hardware/algorithm co-design for ultra-low-bit LLM
weight quantization that enables efficient runtime dequantization without
Hadamard-based incoherence processing. The XOR-Trellis dequantizer combines
shallow parity/XOR logic with structured local reconstruction palettes to
provide highly parallel, ultra-low-complexity reconstruction while preserving
effective trellis search choices. On the quantization side, we reformulate
discrete trellis path optimization with a curvature-aware objective that
reflects model sensitivity in the original coordinate space rather than
transforming the representation for effective Euclidean path selection.
Together, these techniques enable high-dimensional ultra-low-bit trellis
quantization with accurate quantization, inexpensive high-throughput
dequantization, and Hadamard-free deployment.


\bibliography{references}
\bibliographystyle{iclr2027_conference}
\newpage
\appendix
\section{Related Work}\label{app:related_works}

\paragraph{Extreme low-bit PTQ.}
GPTQ uses approximate second-order information for one-shot weight quantization \citep{frantar2022gptq}.
QuIP combines scalar adaptive rounding with incoherence processing \citep{chee2024quip}, while QuIP\# introduces randomized Hadamard transforms and $E_8$ lattice vector quantization \citep{tseng2024quipsharp}.
AQLM represents weight vectors as sums of learned codewords, achieving strong accuracy at very low bitrates \citep{egiazarian2024aqlm}. AQLM uses additive multi-codebook quantization for extreme LLM compression
\citep{egiazarian2024aqlm}, while VPTQ combines vector quantization with
second-order optimization for extreme-low-bit PTQ \citep{liu2024vptq}.
PV-Tuning further improves extreme compression through representation-agnostic
optimization of compressed parameters \citep{malinovskii2024pvtuning};
related second-order model-compression methods include LLM Surgeon
\citep{vanderouderaa2024llmsurgeon}.

\paragraph{Structured codebooks and trellis quantization.}
PVQ uses a pyramid lattice to encode vector directions without storing an explicit codebook \citep{vanderouderaa2024pvq}.
LLVQ uses the 24-dimensional Leech lattice, combining algebraic indexing and angular search with parallel dequantization \citep{vanderouderaa2026llvq}.
Classical trellis-coded quantization (TCQ) and its vector extension couple reconstruction decisions through trellis paths and optimize additive distortion using Viterbi search \citep{marcellin1990tcq,fischer1991tcvq}.
QTIP adapts trellis quantization to LLMs using incoherence processing, a bitshift trellis supporting parallel reconstruction, and computed or hybrid lookup codes \citep{tseng2024qtip}.
Our work builds on this trellis framework, modifying the reconstruction mapping and the path-selection metric.

\paragraph{Curvature-aware objectives and adaptive rounding.}
GPTQ and LDLQ use second-order information to compensate quantization errors through sequential weight updates \citep{frantar2022gptq,chee2024quip}.
GPTVQ uses curvature-weighted distances for codebook fitting and vector assignment \citep{vanbaalen2024gptvq}.
GuidedQuant incorporates end-loss gradients into layerwise objectives and demonstrates improvements with QTIP \citep{kim2025guidedquant}.
YAQA develops Kronecker-factored Hessian sketches and adaptive rounding with both input- and output-side feedback to better preserve end-to-end model behavior \citep{tseng2026yaqa}.
Our trellis search retains the underlying Hessian approximation and error-feedback framework, and uses the diagonal entries of the block-diagonal $D$ factor to weight the additive Viterbi distortion.
This diagonal approximation incorporates residual curvature into path selection while preserving the trellis state space and inference-time reconstruction.

\paragraph{Rotations and Hadamard-free quantization.}
QuaRot uses randomized Hadamard rotations to reduce outliers while preserving the full-precision model function, enabling quantization of weights, activations, and the KV cache \citep{ashkboos2024quarot}.
SpinQuant learns rotations to improve quantized model accuracy, distinguishing rotations absorbed into weights from optional online Hadamard transforms \citep{liu2025spinquant}.
For weight-only PTQ, LLVQ reports competitive Hadamard-free results using shape--gain quantization with norm-preserving GPTQ corrections, interpreted as Spherical GPTQ \citep{vanderouderaa2026llvq}.
Our experiments investigate Hadamard-free weight-only trellis quantization through the combination of the XOR-Trellis reconstruction mapping and curvature-aware path selection.

\paragraph{Efficient reconstruction and low-bit formats.}
SeedLM reconstructs weight blocks from pseudorandom matrices generated from stored seeds and a small set of coefficients, providing another approach to computed reconstruction \citep{shafipour2025seedlm}.
The OCP microscaling specification combines low-bit elements with shared block scales; standard MXFP4 uses FP4 (E2M1) elements and an E8M0 scale per block of 32 elements \citep{ocp2023mx}
Our XOR-Trellis generator maps trellis-state bits to E2M1 payloads using XOR logic and a small lookup, targeting inexpensive fixed-function reconstruction.
The evaluated representation pairs these payloads with E5M3 block scales, so it shares the FP4 element format with MXFP4 but uses a different scale format.
We focus on group-wise FP4 because of its native hardware support and high
low-precision matrix-multiplication throughput in current and next-generation
accelerators, including GPUs. NVIDIA Blackwell provides hardware-accelerated
FP4 Tensor Core support, while the emerging Rubin software stack extends
block-scaled FP4 GEMM support with E5M3 scale factors \citep{nvidia_nvfp4_blackwell,nvidia_cutlass_rubin}. Motivated by this broader
hardware trend toward fine-grained FP4 scaling, our evaluation uses group-wise
FP4 reconstruction with E5M3 scale factors.
\section{Additional Results}\label{app:add_result}
Tables~\ref{tab:llama32_combined_results}--\ref{tab:mistral_7b_results} extend the comparison with QTIP to Llama-3.2, Llama-2, Qwen3, and Mistral models. Across these additional settings, directly removing the Hadamard transform consistently degrades the XOR-Trellis baseline, while diagonal-$D$ weighting substantially recovers the lost performance without introducing any rotation. Energy ordering provides further gains on several models, achieving the strongest rotation-free perplexity and/or average downstream accuracy in many of the evaluated configurations.

On Llama-3.2-1B-Instruct, diagonal-$D$ weighting reduces WT2/C4 perplexity from 30.04/36.35 to 23.04/28.86, while energy ordering further improves it to 21.38/27.16 and raises average zero-shot accuracy from 46.74\% to 51.43\%. A similar trend appears on Llama-3.2-3B-Instruct, where the rotation-free baseline improves from 23.14/23.81 to 15.98/18.86 with diagonal-$D$ weighting. On Llama-2-7B and Llama-2-13B, the Hessian-aware variants likewise recover most or all of the degradation caused by removing RHT, with the best rotation-free configurations outperforming the corresponding QTIP-3INST baselines on perplexity and average downstream accuracy.

The same behavior generalizes beyond the Llama family. On Qwen3-8B, the rotation-free baseline improves from 11.57/15.34 PPL and 63.07\% average accuracy to 10.25/14.19 and 67.75\% with diagonal-$D$ weighting and energy ordering. On Mistral-7B, the corresponding configuration reaches 5.67/8.37 PPL and 67.95\% average accuracy, compared with 6.71/9.78 and 64.39\% for the unweighted rotation-free baseline. Overall, these results reinforce that the benefits of Hessian-aware trellis search are not specific to a single architecture or model scale.
\begin{table}[h]
\centering
\caption{Weight-only PTQ results on Llama-3.2-1B-Instruct and Llama-3.2-3B-Instruct. Perplexity (PPL) is evaluated on WikiText-2 (WT2) and C4 with a context length of 2048, and zero-shot raw accuracy (\texttt{acc}, \%) is reported on five downstream tasks. Lower PPL and higher accuracy indicate better performance.}
\label{tab:llama32_combined_results}
\resizebox{\columnwidth}{!}{%
\begin{tabular}{lllcc|cccccc}
\toprule
& & & \multicolumn{2}{c|}{PPL $\downarrow$}
& \multicolumn{6}{c}{0-Shot Raw Accuracy (\%) $\uparrow$} \\
\cmidrule(lr){4-5}
\cmidrule(lr){6-11}
Model & Method & Rotation & WT2 & C4 & ArcC & ArcE & BoolQ & HSwag & PiQA & Avg. \\
\midrule

\multirow{6}{*}{Llama-3.2-1B-Instruct}
& BF16
& --
& 13.16 & 18.58
& 35.67 & 68.81 & 69.39 & 45.11 & 74.05 & 58.61 \\

& QTIP-3INST
& Hadamard
& 24.28 & 30.44
& 26.19 & 55.56 & 62.23 & 36.32 & 67.08 & 49.48 \\

& XOR-Trellis
& Hadamard
& 22.89 & 28.95
& 25.77 & 56.14 & 60.67 & 37.02 & 66.43 & 49.21 \\

& XOR-Trellis
& None
& 30.04 & 36.35
& 23.89 & 48.44 & \textbf{63.21} & 34.74 & 63.44 & 46.74 \\

& XOR-Trellis + D-weighted
& None
& 23.04 & 28.86
& 26.28 & \textbf{57.11} & 63.00 & 38.09 & 67.57 & 50.41 \\

& XOR-Trellis + D-weighted + energy ordering
& None
& \textbf{21.38} & \textbf{27.16}
& \textbf{29.78} & 56.94 & 62.54 & \textbf{38.56} & \textbf{69.31} & \textbf{51.43} \\

\midrule

\multirow{6}{*}{Llama-3.2-3B-Instruct}
& BF16
& --
& 11.05 & 14.54
& 43.86 & 74.20 & 78.53 & 52.12 & 75.73 & 64.89 \\

& QTIP-3INST
& Hadamard
& 19.30 & 20.91
& 32.25 & 61.07 & 76.54 & 43.39 & 68.06 & 56.26 \\

& XOR-Trellis
& Hadamard
& 16.65 & 19.69
& 33.87 & 64.73 & 73.79 & 44.77 & 71.44 & 57.72 \\

& XOR-Trellis
& None
& 23.14 & 23.81
& 28.75 & 59.09 & 71.47 & 42.54 & 68.28 & 54.03 \\

& XOR-Trellis + D-weighted
& None
& \textbf{15.98} & \textbf{18.86}
& \textbf{33.36} & 65.91 & \textbf{75.57} & \textbf{45.26} & \textbf{70.78} & \textbf{58.18} \\

& XOR-Trellis + D-weighted + energy ordering
& None
& 16.88 & 19.38
& 32.76 & \textbf{65.99} & 73.06 & 44.97 & 70.57 & 57.47 \\

\bottomrule
\end{tabular}%
}
\end{table}

\begin{table}[h]
\centering
\caption{Weight-only PTQ results on Llama-2-7B and Llama-2-13B. Perplexity (PPL) is evaluated on WikiText-2 (WT2) and C4 with a context length of 4096, and zero-shot raw accuracy (\texttt{acc}, \%) is reported on five downstream tasks. Lower PPL and higher accuracy indicate better performance.}
\label{tab:llama2_combined_results}
\resizebox{\columnwidth}{!}{%
\begin{tabular}{lllcc|cccccc}
\toprule
& & & \multicolumn{2}{c|}{PPL $\downarrow$}
& \multicolumn{6}{c}{0-Shot Raw Accuracy (\%) $\uparrow$} \\
\cmidrule(lr){4-5}
\cmidrule(lr){6-11}
Model & Method & Rotation & WT2 & C4 & ArcC & ArcE & BoolQ & HSwag & PiQA & Avg. \\
\midrule

\multirow{6}{*}{Llama-2-7B}
& BF16
& --
& 5.12 & 6.63
& 43.43 & 76.35 & 77.74 & 57.11 & 78.07 & 66.54 \\

& QTIP-3INST
& Hadamard
& 6.95 & 9.08
& 34.56 & 69.53 & 71.93 & 48.15 & 74.48 & 59.73 \\

& XOR-Trellis
& Hadamard
& 6.65 & 8.75
& 36.52 & 71.46 & 70.34 & 49.83 & 73.72 & 60.37 \\

& XOR-Trellis
& None
& 7.34 & 10.07
& 25.17 & 53.87 & 68.41 & 35.87 & 63.38 & 49.34 \\

& XOR-Trellis + D-weighted
& None
& \textbf{6.41} & \textbf{8.52}
& \textbf{37.12} & \textbf{71.17} & \textbf{75.05} & \textbf{51.16} & 74.27 & \textbf{61.75} \\

& XOR-Trellis + D-weighted + energy ordering
& None
& 6.45 & 8.58
& 35.92 & 69.61 & 69.88 & 50.82 & \textbf{74.92} & 60.23 \\

\midrule

\multirow{6}{*}{Llama-2-13B}
& BF16
& --
& 4.57 & 6.05
& 48.46 & 79.34 & 80.70 & 60.05 & 79.16 & 69.54 \\

& QTIP-3INST
& Hadamard
& 5.64 & 7.59
& 41.47 & 76.30 & 79.60 & 53.69 & 76.50 & 65.51 \\

& XOR-Trellis
& Hadamard
& 5.55 & 7.30
& 40.36 & 75.59 & 80.21 & 54.77 & 76.93 & 65.57 \\

& XOR-Trellis
& None
& 5.94 & 7.96
& 40.19 & 73.78 & 78.72 & 52.68 & 74.97 & 64.07 \\

& XOR-Trellis + D-weighted
& None
& \textbf{5.47} & 7.36
& 41.04 & 76.05 & 78.56 & 54.83 & 76.44 & 65.38 \\

& XOR-Trellis + D-weighted + energy ordering
& None
& 5.48 & \textbf{7.34}
& \textbf{43.60} & \textbf{76.89} & \textbf{79.79} & \textbf{55.51} & \textbf{77.15} & \textbf{66.59} \\

\bottomrule
\end{tabular}%
}
\end{table}

\begin{table}[h]
\centering
\caption{Weight-only PTQ results on Qwen3-8B. Perplexity (PPL) is evaluated on WikiText-2 (WT2) and C4 with a context length of 4096, and zero-shot raw accuracy (\texttt{acc}, \%) is reported on five downstream tasks. Lower PPL and higher accuracy indicate better performance.}
\label{tab:qwen3_8b_results}
\resizebox{\columnwidth}{!}{%
\begin{tabular}{llcc|cccccc}
\toprule
& & \multicolumn{2}{c|}{PPL $\downarrow$}
& \multicolumn{6}{c}{0-Shot Accuracy (\%) $\uparrow$} \\
\cmidrule(lr){3-4}
\cmidrule(lr){5-10}
Method & Rotation & WT2 & C4 & ArcC & ArcE & BoolQ & HSwag & PiQA & Avg. \\
\midrule
BF16
    & --
    & 9.00 & 12.48
    & 55.46 & 83.54 & 86.54 & 57.04 & 76.55 & 71.83 \\

QTIP-3INST
    & Hadamard
    & 11.35 & 14.97
    & 48.63 & 78.49 & 85.90 & 49.93 & 74.65 & 67.52 \\

XOR-Trellis
    & Hadamard
    & 10.29 & 14.23
    & 47.70 & 78.28 & 84.01 & 51.30 & 74.65 & 67.19 \\

\midrule
XOR-Trellis
    & None
    & 11.57 & 15.34
    & 40.61 & 70.12 & 82.54 & 50.33 & 71.76 & 63.07 \\

XOR-Trellis + D-weighted
    & None
    & 10.34 & 14.38
    & 46.93 & 78.07 & 83.70 & 51.09 & \textbf{74.59} & 66.88 \\

XOR-Trellis + D-weighted + energy ordering
    & None
    & \textbf{10.25} & \textbf{14.19}
    & \textbf{48.21} & \textbf{79.04} & \textbf{85.41} & \textbf{51.75} & 74.32 & \textbf{67.75} \\
\bottomrule
\end{tabular}%
}
\end{table}

\begin{table}[h]
\centering
\caption{Weight-only PTQ results on Mistral-7B. Perplexity (PPL) is evaluated on WikiText-2 (WT2) and C4 with a context length of 4096, and zero-shot raw accuracy (\texttt{acc}, \%) is reported on five downstream tasks. Lower PPL and higher accuracy indicate better performance.}
\label{tab:mistral_7b_results}
\resizebox{\columnwidth}{!}{%
\begin{tabular}{llcc|cccccc}
\toprule
& & \multicolumn{2}{c|}{PPL $\downarrow$}
& \multicolumn{6}{c}{0-Shot Accuracy (\%) $\uparrow$} \\
\cmidrule(lr){3-4}
\cmidrule(lr){5-10}
Method & Rotation & WT2 & C4 & ArcC & ArcE & BoolQ & HSwag & PiQA & Avg. \\
\midrule
BF16
    & --
    & 4.91 & 7.42
    & 50.17 & 80.81 & 83.73 & 61.25 & 80.52 & 71.30 \\

QTIP-3INST
    & Hadamard
    & 6.04 & 8.77
    & 42.75 & 75.67 & 82.72 & 54.62 & 78.24 & 66.80 \\

XOR-Trellis
    & Hadamard
    & 5.68 & 8.41
    & 43.60 & 76.81 & 83.06 & 55.99 & 79.22 & 67.74 \\

\midrule
XOR-Trellis
    & None
    & 6.71 & 9.78
    & 38.74 & 73.11 & 80.64 & 52.86 & 76.61 & 64.39 \\

XOR-Trellis + D-weighted
    & None
    & 5.71 & 8.41
    & \textbf{44.88} & 76.73 & 82.63 & 56.17 & 78.62 & 67.81 \\

XOR-Trellis + D-weighted + energy ordering
    & None
    & \textbf{5.67} & \textbf{8.37}
    & 44.45 & \textbf{77.31} & \textbf{82.81} & \textbf{56.45} & \textbf{78.73} & \textbf{67.95} \\
\bottomrule
\end{tabular}%
}
\end{table}

\providecommand{\danisotropyfigdir}{figures/d_anisotropy}

\section{Theory of the XOR-Trellis Generation Hash}
\label{app:generation-hash-theory}

This section gives a more formal view of the finite-state generation function used in Section~\ref{sec:xor-decoder}.  The central observation is that a trellis generator has three logically distinct design objects: (i) the \emph{global reproduction law}, which controls how frequently numerical symbols occur over the generated state space; (ii) the \emph{local palette geometry}, which controls which reconstruction values are simultaneously available from a shared history; and (iii) the \emph{state/hash topology}, which controls how palettes and branch labels evolve along a path.  Matching the first object alone is therefore insufficient: two generators can have the same aggregate reconstruction histogram but expose different local menus to Viterbi or connect those menus with different temporal structure.  This distinction is classical in spirit---trellis source coding is a finite-state minimum-distortion path problem \citep{gray1977trellis,marcellin1990tcq,fischer1991tcvq}---but it is particularly useful for understanding why the structured palettes in Section~\ref{sec:xor-decoder} matter.

\subsection{Finite-state generation model}\label{app:finite-state}
Let $h_t\in\mathbb F_2^{\ell}$ denote the retained trellis history and $b_t\in\mathbb F_2^2$ the two-bit branch.  The XOR-Trellis construction can be written abstractly as
\begin{align}
  h_{t+1} &= S h_t + B b_t, \\
  c_t &= A h_t \in \mathbb F_2^2, \\
  u_t &= b_t + Q h_t \in \mathbb F_2^2, \\
  \hat x_t &= P_{c_t}[u_t],
  \label{eq:hash-linear-model}
\end{align}
where additions are over $\mathbb F_2$.  The two-bit value $c_t$ selects one of four palettes and $u_t$ selects a slot within that palette.  For a fixed history $h$, the local menu presented to Viterbi is
\begin{equation}
  \mathcal P(h)=\{g(h,b):b\in\mathbb F_2^2\}.
\end{equation}
If uniform states and branches are used only as a design measure, the aggregate candidate marginal $q_g(a)$ counts the fraction of state/branch pairs that generate symbol $a$.  Importantly, this is not generally the empirical marginal of the Viterbi-selected path, because path selection depends on the source.

Rate--distortion theory motivates matching a useful aggregate reproduction law in the long-block unconstrained limit \citep{shannon1959coding,blahut1972rate}.  It does not specify which four symbols should be siblings at one state, nor how those palettes should be connected by a finite-memory trellis.  Finite memory therefore creates a separate local-design problem: in our two-bit construction, every state exposes only four immediate choices regardless of the total number of trellis states.

\subsection{Why the four XOR-Trellis palettes are locally optimal}
\label{app:palette-optimality}
Consider the 16 labeled FP4 payloads in sorted order,
\begin{equation}
\{-6,-4,-3,-2,-1.5,-1,-0.5,0^{-},0^{+},0.5,1,1.5,2,3,4,6\},
\end{equation}
where the two numerical zeros are retained as distinct payload labels.  The palettes used by XOR-Trellis are obtained by assigning sorted ranks modulo four:
\begin{align}
P_0&=\{-6,-1.5,0^{+},2\}, &
P_1&=\{-4,-1,0.5,3\},\\
P_2&=\{-3,-0.5,1,4\}, &
P_3&=\{-2,0^{-},1.5,6\}.
\end{align}
This construction admits a simple pointwise optimality result for the immediate four-way choice.

\paragraph{Proposition 1 (pointwise one-step optimality).}
Partition the 16 labeled payloads into four disjoint four-element palettes $P_0,\ldots,P_3$.  Assume the palette identity is uniform and independent of a scalar source value $x$, and define
\begin{equation}
D_{\mathcal P}(x)=\frac14\sum_{c=0}^{3}\min_{a\in P_c}(x-a)^2.
\end{equation}
Let $d_{(1)}(x)\le\cdots\le d_{(16)}(x)$ be the ordered squared distances from $x$ to the 16 labeled payloads.  Any partition satisfies
\begin{equation}
D_{\mathcal P}(x)\ge \frac14\sum_{j=1}^{4}d_{(j)}(x),
\end{equation}
and the rank-modulo-four partition above attains equality for every $x\in\mathbb R$.

\paragraph{Proof sketch.}
The four palette minima must correspond to four distinct labeled payloads, so their total distance cannot be smaller than the distance to the four globally nearest labels.  In one dimension, the four nearest members of a sorted alphabet can be chosen as four consecutive ranks (including at ties).  Four consecutive ranks have four different residues modulo four; hence the rank-interleaved partition places one of those four nearest labels in every palette and attains the lower bound. \hfill$\square$

Because the equality is pointwise, the same palettes minimize the expected one-step distortion for \emph{any} scalar source distribution for which the expectation exists, under the stated fixed-inventory and uniform-palette assumptions.  The result is deliberately local: it does not establish optimality for a source-correlated palette process, a Hessian-weighted/vector distortion, or the full multi-step Viterbi path objective.

A second useful property concerns numerical separation.  Among all partitions of these 16 labeled payloads into four palettes, the largest achievable minimum within-palette numerical spacing is 1.5, and the rank-interleaved palettes attain it.  Five central labels lie within an interval of width 1.5, so the pigeonhole principle prevents any four-way partition from achieving larger minimum spacing; direct inspection shows that our palettes meet the bound.  Thus no state exposes two nearly identical reconstruction choices.

The local argument also explains why random symbol assignment can waste branches.  If four ordered reproduction strata each carry probability $1/4$ and four outgoing symbols are drawn independently, the probability that all four strata appear is only
\begin{equation}
  \frac{4!}{4^4}=\frac{3}{32}=9.375\%.
\end{equation}
A stratified palette construction can preserve the same global inventory while forcing broad local coverage.  This calculation is an IID-menu diagnostic, not a proof of optimal full-trellis distortion.

\subsection{Linear-rank theory for the XOR mapping}
\label{app:xor-rank-theory}
The palette construction fixes local geometry; the XOR masks determine how those palettes evolve along paths.  Because the mapping is linear over $\mathbb F_2$, several structural defects can be detected exactly with rank tests.  If $h$ is uniform and $\operatorname{rank}_2(A)=2$, then $c=Ah$ is uniform over the four palettes.  For current and next palette labels,
\begin{equation}
\begin{bmatrix}c_t\\c_{t+1}\end{bmatrix}
=
\underbrace{\begin{bmatrix}A&0\\AS&AB\end{bmatrix}}_{E_2}
\begin{bmatrix}h_t\\b_t\end{bmatrix}.
\end{equation}
All 16 adjacent palette pairs are equally represented under uniform design inputs if and only if
\begin{equation}
  \operatorname{rank}_2(E_2)=4.
\end{equation}
More generally, stacking $w$ consecutive palette labels yields a binary observability/drive matrix $E_w$; full rank $\operatorname{rank}_2(E_w)=2w$ certifies uniform support over all $4^w$ palette words under the same design measure.  Rank deficits therefore expose deterministic parity constraints that may be invisible in a balanced one-step histogram.

These tests are best interpreted as \emph{rejection criteria}, not as an optimality theorem for a particular XOR mask.  Full rank certifies support/entropy properties under uniform branch bits, but does not order the Euclidean distortion of Viterbi-selected paths.  A practical design procedure is therefore: preserve the desired symbol inventory, construct locally diverse palettes, reject XOR masks with short-window rank defects, and choose among the surviving masks using held-out full-path distortion at the actual trellis memory and coding horizon.

\subsection{Implications for the XOR-Trellis dequantizer design}
The theory suggests a hierarchy that mirrors the empirical design in Section~\ref{sec:xor-decoder}.  First choose the source/distortion and a useful global reproduction inventory; next partition that inventory into locally useful palettes; then assign palettes and branch permutations to states with a cheap topology that passes structural rank checks; finally evaluate full Viterbi distortion under the actual block length, boundary rule, scales, and Hessian-aware metric.  The key consequence for this paper is that the XOR-Trellis generator is not justified by its aggregate FP4 histogram alone.  Its four palettes have an exact one-step covering property under explicit assumptions, while the XOR hash provides a low-cost topology whose obvious degeneracies can be screened algebraically.  Full finite-horizon performance remains an operational property and is evaluated empirically.

\section{Theory of Curvature-Aware Trellis Path Optimization}
\label{app:hessian-viterbi-theory}

This section formalizes the objective-mismatch view developed in Section~\ref{sec:hessian-search}. The key question is not merely whether a diagonal weight can improve an existing Viterbi metric, but when isotropic discrete path selection is a faithful surrogate for the second-order geometry that motivates the surrounding PTQ procedure. We use the $LDL^\top$ factorization to expose the residual quadratic metric, quantify the loss of replacing that metric by an identity weighting, and show that curvature-aware branch costs preserve the standard Viterbi dynamic program. The analysis does not claim that $D$ alone represents the full Hessian or that diagonal weighting subsumes the effects of RHT.

\subsection{The LDL quadratic form}
For a reconstruction error vector $e=w-\hat w$ and a positive-definite local Hessian approximation
\begin{equation}
    H=LDL^\top,
\end{equation}
the second-order proxy can be rewritten exactly as
\begin{align}
    e^\top H e
    &= e^\top L D L^\top e \\
    &= (L^\top e)^\top D(L^\top e).
\end{align}
Defining the transformed residual
\begin{equation}
    r=L^\top e,
    \label{eq:transformed-residual}
\end{equation}
gives
\begin{equation}
    \boxed{e^\top H e=r^\top D r=\sum_i D_{ii}r_i^2.}
    \label{eq:ldl-diagonalized-loss}
\end{equation}
Thus the off-diagonal coupling of the original quadratic form is represented by the triangular change of coordinates $L^\top$, while the remaining curvature in those coordinates is diagonal.  This identity is the formal basis for the intuition in Section~\ref{sec:hessian-search}: $L$ captures cross-coordinate interaction, whereas $D$ measures the relative cost of residual error after that interaction is represented.

The LDLQ implementation operates sequentially on feedback-adjusted targets rather than explicitly forming the full vector $r$ in Equation~\ref{eq:transformed-residual}.  Consequently, we do not require the identity $r_i=z_i-\hat z_i$ as an additional assumption.  Instead, our proposed trellis metric uses the diagonal curvature exposed by the same factorization to weight the residuals presented to the discrete search:
\begin{equation}
    \mathcal L_D(\pi)=\sum_i D_{ii}\,\rho_i(\pi)^2,
    \qquad
    \rho_i(\pi)=z_i-\hat z_i(\pi).
    \label{eq:appendix-diagD-path}
\end{equation}
Equation~\ref{eq:ldl-diagonalized-loss} therefore motivates $D$ as the natural per-coordinate sensitivity signal, while Equation~\ref{eq:appendix-diagD-path} is the operational additive objective optimized by our Viterbi search.  Establishing an exact coordinate-by-coordinate equality between $\rho$ and $L^\top e$ would depend on the particular LDLQ feedback convention; our method does not rely on such an equality.

\subsection{Euclidean Viterbi as an isotropic approximation}
For any residual vector $r$, let
\begin{equation}
    d_{\min}=\min_i D_{ii},\qquad
    d_{\max}=\max_i D_{ii},\qquad
    \kappa_D=\frac{d_{\max}}{d_{\min}}.
\end{equation}
Because $D$ is diagonal and positive,
\begin{equation}
    d_{\min}\lVert r\rVert_2^2
    \le r^\top D r
    \le d_{\max}\lVert r\rVert_2^2.
    \label{eq:diagD-norm-equivalence}
\end{equation}
Hence replacing $D$ by the identity amounts to an isotropic approximation to the residual quadratic metric.  When $D\approx \alpha I$, or equivalently $\kappa_D\approx1$, the two objectives differ primarily by a positive scale factor and therefore induce nearly the same ordering of candidate residuals.  As the spread of the diagonal entries grows, this surrogate can become progressively less faithful.

The same observation yields a simple path-level approximation bound for the diagonal objective.  Let
\begin{align}
    \pi_E &\in \arg\min_{\pi}\lVert r(\pi)\rVert_2^2,\\
    \pi_D &\in \arg\min_{\pi}r(\pi)^\top D r(\pi).
\end{align}
Then
\begin{align}
    \mathcal L_D(\pi_E)
    &\le d_{\max}\lVert r(\pi_E)\rVert_2^2 \\
    &\le d_{\max}\lVert r(\pi_D)\rVert_2^2 \\
    &\le \kappa_D\,\mathcal L_D(\pi_D),
\end{align}
so
\begin{equation}
    \boxed{\mathcal L_D(\pi_E)\le \kappa_D\,\mathcal L_D(\pi_D).}
    \label{eq:kappa-approx-bound}
\end{equation}
This bound is deliberately coarse: it uses the extreme diagonal entries and ignores trellis structure.  Its value is conceptual.  It makes explicit that Euclidean path search is well justified when the residual diagonal metric is nearly isotropic, whereas the guarantee weakens as diagonal anisotropy increases.  In the empirical analysis of Appendix~\ref{app:d-anisotropy}, we use the more robust statistic $P_{95}(D_{ii})/P_5(D_{ii})$ rather than $\kappa_D$ to avoid domination by extreme entries.

\subsection{Connection to RHT}
Let $R$ denote an orthogonal incoherence transform.  In the rotated coordinates the quadratic form becomes
\begin{equation}
    e_R^\top H_R e_R,
    \qquad
    H_R=R H R^\top,
    \qquad
    e_R=R e.
\end{equation}
RHT does not remove curvature; it changes the basis in which that curvature and the weight distribution are presented to the quantizer.  If this mixing makes the residual sensitivities substantially more homogeneous, then an unweighted branch metric becomes a better local surrogate.  Our no-RHT method takes the complementary route: it keeps the original representation and retains explicit diagonal sensitivity in the branch cost.  The two procedures are therefore related but not equivalent.  In particular, a diagonal metric cannot represent residual off-diagonal coupling that remains in the original basis, which explains why diagonal anisotropy alone need not predict whether no-RHT $D$-weighting beats a complete RHT pipeline on every matrix.

\subsection{The weighted objective preserves Viterbi optimality and complexity}
Let $s_t$ be the trellis state, $b_t$ a branch, $f(s_t,b_t)$ the next-state function, and $g(s_t,b_t)$ the reconstruction generated by that transition.  For target $z_t$, define the proposed branch cost
\begin{equation}
    c_t(s_t,b_t)
    =D_{tt}\bigl(z_t-g(s_t,b_t)\bigr)^2.
    \label{eq:weighted-branch-cost}
\end{equation}
The path cost remains additive,
\begin{equation}
    C(\pi)=\sum_{t=1}^{T}c_t(s_t,b_t),
\end{equation}
and therefore obeys the standard Viterbi/Bellman recurrence
\begin{equation}
    V_{t+1}(s')=
    \min_{\substack{s,b:\\ f(s,b)=s'}}
    \left[V_t(s)+D_{tt}\bigl(z_t-g(s,b)\bigr)^2\right].
    \label{eq:weighted-viterbi-recurrence}
\end{equation}
Thus $D$-weighting changes only the scalar branch score.  It does not increase the number of trellis states, outgoing transitions, survivor paths, or the asymptotic dynamic-programming complexity, and it does not change the compressed bitrate.

\subsection{Offline-only cost}
The additional branch-score operation is a multiplication by the already available scalar $D_{tt}$ after the squared residual has been formed.  This work occurs only during offline quantization.  Neither $D$ nor any Hessian-derived quantity is stored in the deployed trellis bitstream, and the runtime decoder of Section~\ref{sec:xor-decoder} is unchanged.  The mathematical refinement of the search objective therefore remains cleanly separated from the ultra-low-complexity inference-time reconstruction path.

\section{Mechanistic Analysis of Hessian Anisotropy}
\label{app:d-anisotropy}

This section studies why diagonal-$D$-weighted Viterbi decoding is especially
effective when randomized Hadamard transforms (RHTs) are disabled. Two related
effects arise in the unrotated domain. First, groupwise scaling remains
sensitive to within-group outliers, which can force most weights into a narrow
region around zero. Second, the local Hessian metric can be strongly
anisotropic, so equal Euclidean branch-cost weighting does not reflect the
unequal importance of different coordinates. RHT can mitigate both effects by
redistributing weight magnitudes and making the local metric more isotropic,
whereas our method retains the original basis and incorporates the diagonal of
the block metric directly into the Viterbi branch cost. 
\subsection{Why Groupwise Scaling Does Not Eliminate the Need for RHT}
\label{app:groupwise_scaling}

Groupwise scaling mitigates dynamic-range variation, but it does not fully
remove the difficulty caused by outliers in the unrotated domain. A group may
contain one or more weights whose magnitudes are substantially larger than
those of most other weights in the same group. Because a shared scale must
accommodate these large values within the available FP4 reconstruction range,
the outliers can dominate the scale selection. As a result, the majority of
smaller-magnitude weights are compressed into a relatively narrow region around
zero.

In our analysis of RHT-sensitive matrices, a substantial fraction of the
quantization error is concentrated in this near-zero region. This suggests that
groupwise scaling alone does not eliminate the conditioning problem addressed
by RHT: it controls the range within each group, but does not redistribute
outlier energy across coordinates. RHT can reduce this imbalance by mixing
coordinates before quantization, while the rotation-free method must instead
remain robust to the resulting non-uniform geometry in the original basis.

This effect is complementary to the Hessian anisotropy studied below. Scale
distortion describes how the reconstruction grid is used within a group,
whereas diagonal-$D$ anisotropy describes how costly errors are along different
coordinates. The two mechanisms therefore act at different parts of the
quantization problem.

\subsection{Definitions and protocol}

We next isolate the second mechanism---anisotropy in the local Hessian metric.
For each weight matrix $W$, Hessian $H$, and reconstruction $\widehat W$, we
measure normalized full-Hessian error

\begin{equation}
  E(\widehat W;W,H)
  = \frac{\operatorname{tr}\!\left[(\widehat W-W)H
      (\widehat W-W)^\top\right]}
    {\operatorname{tr}\!\left[WHW^\top\right]}.
  \label{eq:relative-full-hessian-error}
\end{equation}
This normalization makes errors comparable across matrices with different
dimensions and weight energies.  We use QTIP's block factorization
$H=LDL^\top$ with 16-column blocks.  Legacy decoding uses $L$ for sequential
error feedback but uses an unweighted Euclidean Viterbi branch metric.  The
diagonal-$D$ variant instead uses
\begin{equation}
  c_b(q;t)=\sum_{i\in b} D_{b,ii}(q_i-t_i)^2,
  \label{eq:diag-d-branch-cost}
\end{equation}
where $t$ is the feedback-adjusted target, $q$ is the trellis reconstruction.  This modification represents unequal
conditional feature sensitivity, but it does not model the off-diagonal
entries of $D_b$.

We quantify diagonal anisotropy using the robust statistic
\begin{equation}
  A_D=\frac{P_{95}(D_{ii})}{P_{5}(D_{ii})},
  \qquad x=\log A_D.
  \label{eq:robust-d-anisotropy}
\end{equation}
The analysis covers all seven projections (Q, K, V, O, up, gate, and down) in
all 32 layers of Llama-3.1-8B, for 224 matrices in total.  RHT and no-RHT
conditions use matched weights, Hessians, FP4 reconstruction, scales, and
quantization settings.  Confidence intervals are 95\% percentile intervals
from 2,000 bootstrap samples that resample complete transformer layers, thereby
preserving dependence among projections in the same layer.

\subsection{Anisotropy predicts where metric awareness matters}

Table~\ref{tab:d-anisotropy-correlations} summarizes the matrix-level
relationships.  The no-RHT diagonal-$D$ gain is
\begin{equation}
  G_D=\log\frac{E_{\mathrm{noRHT,legacy}}}
                       {E_{\mathrm{noRHT,diagD}}},
  \label{eq:diag-d-gain}
\end{equation}
and the legacy benefit from RHT is
\begin{equation}
  G_{\mathrm{RHT}}=\log\frac{E_{\mathrm{noRHT,legacy}}}
                            {E_{\mathrm{RHT,legacy}}}.
  \label{eq:rht-gain}
\end{equation}
Positive values denote lower error for the named intervention.

\begin{table*}[t]
  \centering
  \small
  \caption{Spearman correlations with no-RHT robust anisotropy $\log A_D$.
  Intervals use complete-layer bootstrap samples.  The absolute error delta
  and direct no-RHT-diagD comparison are post-hoc diagnostics.}
  \label{tab:d-anisotropy-correlations}
  \begin{tabular}{lrr}
    \toprule
    Matrix-level outcome & $\rho$ & 95\% CI \\
    \midrule
    no-RHT diagonal-$D$ gain $G_D$
      & 0.545 & [0.454, 0.631] \\
    legacy RHT gain $G_{\mathrm{RHT}}$
      & 0.629 & [0.537, 0.718] \\
    absolute legacy error reduction,
      $E_{\mathrm{noRHT}}-E_{\mathrm{RHT}}$
      & 0.606 & [0.535, 0.676] \\
    no-RHT diagD advantage over legacy RHT,
      $\log(E_{\mathrm{RHT,legacy}}/E_{\mathrm{noRHT,diagD}})$
      & $-0.291$ & [$-0.414$, $-0.172$] \\
    \bottomrule
  \end{tabular}
\end{table*}

\begin{figure*}[t]
  \centering
  \begin{minipage}[t]{0.325\textwidth}
    \centering
    \includegraphics[width=\linewidth]{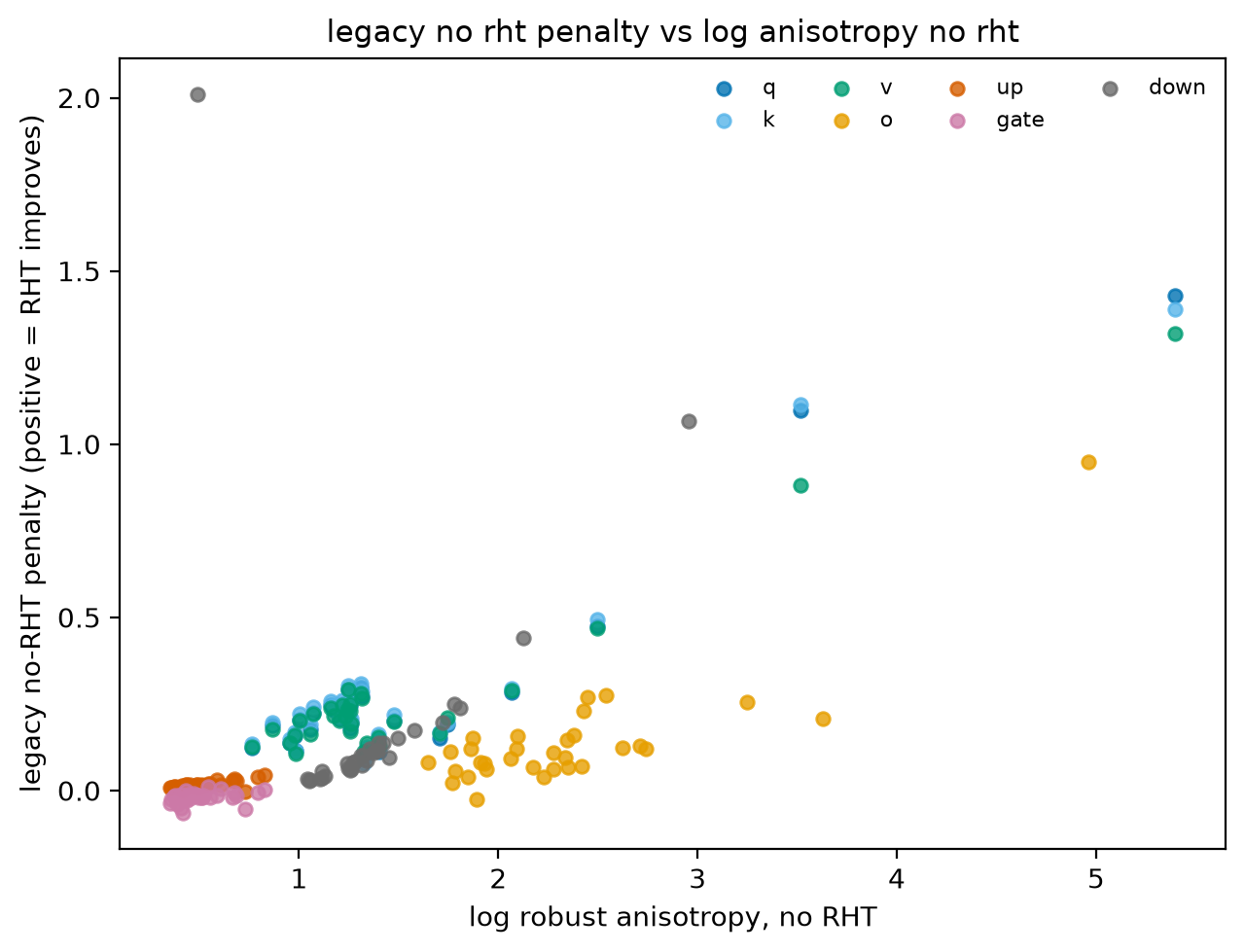}
    \small (a) Legacy RHT log-error gain.
  \end{minipage}\hfill
  \begin{minipage}[t]{0.325\textwidth}
    \centering
    \includegraphics[width=\linewidth]{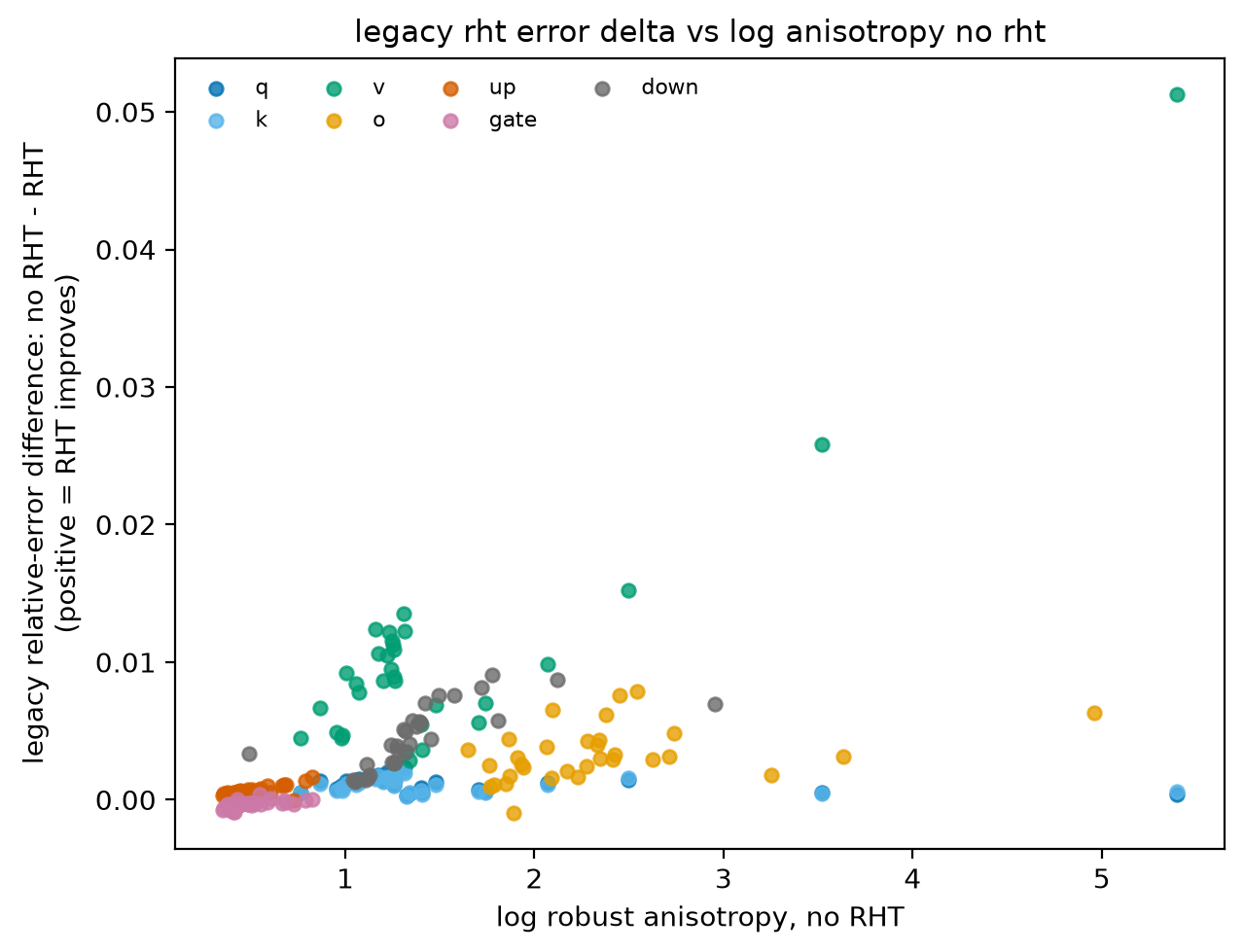}
    \small (b) Legacy RHT absolute error reduction.
  \end{minipage}\hfill
  \begin{minipage}[t]{0.325\textwidth}
    \centering
    \includegraphics[width=\linewidth]{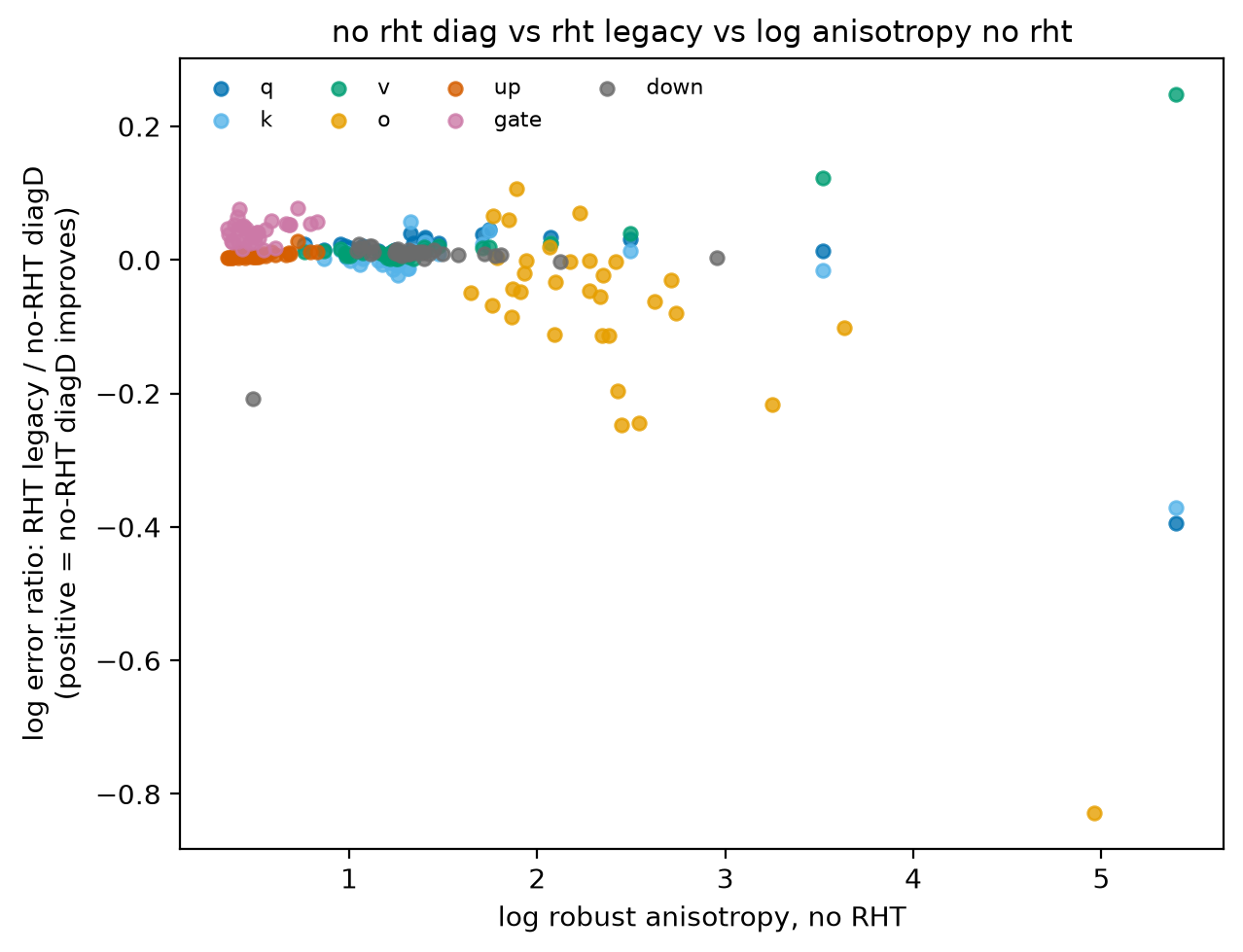}
    \small (c) No-RHT diagD versus legacy RHT.
  \end{minipage}
  \caption{Local full-Hessian error effects versus no-RHT robust diagonal
  anisotropy.  (a) RHT reduces legacy error by a larger factor for more
  anisotropic matrices ($\rho=0.629$).  (b) The corresponding absolute
  relative-error reduction follows the same trend ($\rho=0.606$).  Positive
  values in (a) and (b) favor RHT.  (c) Positive values favor no-RHT
  diagonal-$D$ over legacy RHT.  Its negative association ($\rho=-0.291$)
  shows that diagonal anisotropy alone does not determine which complete
  method wins.}
  \label{fig:d-anisotropy-local-effects}
\end{figure*}

The positive correlation between anisotropy and $G_D$ supports the proposed
mechanism: when conditional feature sensitivities are uneven, assigning equal
branch-cost weight to every coordinate is increasingly mismatched to the
Hessian metric.  The similarly strong relationship with $G_{\mathrm{RHT}}$
shows that RHT is most helpful to legacy QTIP for the same anisotropic
matrices.  Figure~\ref{fig:d-anisotropy-local-effects}(a) visualizes this trend
using a scale-independent log error ratio; each point is one matrix and colors
denote projection type.  Figure~\ref{fig:d-anisotropy-local-effects}(b)
presents the same mechanism using the
absolute difference between normalized errors.  Its positive correlation
shows that the ratio result is not solely an artifact of very small
denominators.  Unlike the log ratio, however, this quantity remains sensitive
to the absolute relative-error scale of each matrix.

\subsection{Why diagonal weighting does not subsume RHT}

Although diagonal weighting and RHT both address anisotropy, they are not
equivalent.  RHT changes the basis and can redistribute both diagonal energy
and cross-coordinate coupling.  Equation~\ref{eq:diag-d-branch-cost} corrects
only the diagonal importance weights in the original basis.  Consequently,
greater anisotropy predicts improvement over the \emph{no-RHT legacy}
baseline, but does not predict that no-RHT diagonal-$D$ decoding will beat
\emph{legacy RHT} for each individual matrix.
Figure~\ref{fig:d-anisotropy-local-effects}(c) makes this distinction explicit
and is consistent with residual off-diagonal structure and projection-specific
effects.

There are informative matrix-level exceptions.  Diagonal-$D$ decoding
improves over legacy no-RHT decoding for all 224 measured matrices, but several
O projections have high anisotropy and only modest gains.  For example,
layer-27 O has $\log A_D=2.45$ and $G_D=0.023$, while layer-31 O has
$\log A_D=3.25$ and $G_D=0.039$.  Layer-0 O is particularly notable:
$\log A_D=4.96$, yet legacy RHT remains substantially better than no-RHT
diagonal-$D$ decoding.  Conversely, layer-1 down obtains an unusually large
diagonal-$D$ gain despite low anisotropy.  These cases show that off-diagonal
correlation, projection structure, and trellis/codebook compatibility explain
variation beyond the diagonal statistic.

\subsection{Full-model consistency and limitations}

The matrix-level mechanism is consistent with the matched WikiText-2 result:
no-RHT diagonal-$D$ decoding achieves perplexity 8.06145, compared with
8.10438 for legacy RHT, 8.68979 for legacy no-RHT, and 8.19951 for RHT with
diagonal-$D$ weighting.  These endpoints show that explicitly representing
anisotropy can recover the accuracy lost when RHT is unavailable.  They also
show that diagonal weighting is not automatically beneficial after RHT: once
the basis has changed, dense within-block structure can matter more than its
diagonal approximation.

This analysis should not be read as evidence that preserved correlation is an
advantage of no-RHT quantization.  Correlation that remains outside the
diagonal model is a residual source of error.  Moreover, matrix-level proxy
correlations do not establish causality, the post-hoc diagnostics were selected
after observing the primary analysis, and a single scalar anisotropy statistic
cannot characterize the full block metric.  The supported conclusion is
narrower: robust diagonal anisotropy is a strong, statistically stable
predictor of both legacy RHT benefit and no-RHT diagonal-weighting benefit, but
it is not a complete predictor of the best quantization method for every
matrix.

\section{XOR-Trellis Generator and Decorrelation Analysis}
\label{app:xor_stratified_decorrelation}

The XOR-Trellis generator maps each 16-bit trellis state to one of the
16 FP4 payload codes.  For a state $x$, the two least-significant bits
define the transition slot $u=x\bmod 4$, while the remaining bits define
the shared history $h=\lfloor x/4\rfloor$.  Four masked parity functions
produce a two-bit palette index and a two-bit, history-dependent
permutation:
\begin{align}
p_i &= \operatorname{parity}(h \mathbin{\&} m_i),
    && i\in\{0,1\},\\
q_i &= \operatorname{parity}(h \mathbin{\&} m_{i+2}),
    && i\in\{0,1\},\\
p &= p_0 + 2p_1,\qquad
j = u \oplus (q_0+2q_1).
\end{align}
The output is the $j$-th entry of palette $\mathcal{P}_p$.  The four
palettes, expressed as reconstruction values, are
\begin{align}
\mathcal{P}_0 &= [-6,-1.5,0,2],&
\mathcal{P}_1 &= [-4,-1,0.5,3],\\
\mathcal{P}_2 &= [-3,-0.5,1,4],&
\mathcal{P}_3 &= [-2,0,1.5,6].
\end{align}
Thus, every shared 14-bit history exposes four distinct reconstruction
values spanning both signs, while the permutation changes the association
between transition labels and palette entries without changing the local
set of choices.

\begin{lstlisting}[language=Python,
                   basicstyle=\ttfamily\footnotesize,
                   caption={Exact XOR-Trellis MXFP4 generator.},
                   label={lst:XOR-Trellis}]
MXFP4_VALUES = torch.tensor([
    -6.0, -4.0, -3.0, -2.0, -1.5, -1.0, -0.5, 0.0,
     0.0,  0.5,  1.0,  1.5,  2.0,  3.0,  4.0, 6.0,
])

PALETTES = torch.tensor([
    [0, 4,  7, 12],  # [-6, -1.5, 0,   2]
    [1, 5,  9, 13],  # [-4, -1,   0.5, 3]
    [2, 6, 10, 14],  # [-3, -0.5, 1,   4]
    [3, 8, 11, 15],  # [-2,  0,   1.5, 6]
], dtype=torch.long)

MASKS = (0x1089, 0x2112, 0x0A24, 0x2441)

def parity_masked(x, mask):
    y = x & mask
    y ^= y >> 8
    y ^= y >> 4
    y ^= y >> 2
    y ^= y >> 1
    return y & 1

def xor_stratified_codes(states):
    x = states.to(torch.int64) & 0xFFFF
    slot = x & 3
    history = x >> 2

    p0 = parity_masked(history, MASKS[0])
    p1 = parity_masked(history, MASKS[1])
    q0 = parity_masked(history, MASKS[2])
    q1 = parity_masked(history, MASKS[3])

    palette = p0 | (p1 << 1)
    entry = slot ^ (q0 | (q1 << 1))
    return PALETTES.to(x.device)[palette, entry]
\end{lstlisting}

\paragraph{Transition decorrelation.}
We exhaustively evaluate all $2^{16}$ states and each of the four possible
$K=2$ state transitions.  For appended transition bits
$r\in\{0,1,2,3\}$, the successor state is
\begin{equation}
T_r(x)=\bigl((x\ll 2)\bmod 2^{16}\bigr)\mathbin{\vert}r.
\end{equation}
Let $g(x)$ denote the signed FP4 reconstruction value produced by the
generator.  We measure signed and magnitude transition correlations as
\begin{align}
\rho_r^{\mathrm{signed}}
    &=\operatorname{Corr}\!\left(g(x),g(T_r(x))\right),\\
\rho_r^{\mathrm{abs}}
    &=\operatorname{Corr}\!\left(|g(x)|,|g(T_r(x))|\right),
\end{align}
where the correlation is computed uniformly over all 65,536 states.
We additionally compute pooled correlations by concatenating the samples
from all four transitions before applying the same Pearson-correlation
formula.

\begin{table}[t]
\centering
\caption{Exhaustive transition-correlation results for XOR-Trellis.
Values are computed over all $2^{16}$ states.}
\label{tab:XOR-Trellis-correlation}
\begin{tabular}{lccccc}
\toprule
Metric & $r=00$ & $r=01$ & $r=10$ & $r=11$ & Pooled\\
\midrule
Signed correlation
    & 0.000000 & 0.000000 & 0.000000 & 0.000000 & 0.000000\\
Absolute-value correlation
    & 0.000000 & 0.000000 & 0.000000 & 0.000000 & 0.000000\\
\bottomrule
\end{tabular}
\end{table}

The zero signed correlation indicates that the sign and numeric value at
the next trellis step are not linearly predictable from the current
output.  The zero absolute-value correlation further shows that large
magnitudes are not systematically followed by large magnitudes, or small
magnitudes by small magnitudes.  This property holds separately for every
possible transition label and after pooling all transitions.  It should
be interpreted as exact zero Pearson correlation under exhaustive state
enumeration, rather than as a claim of statistical independence.
Consequently, the generator combines decorrelated successive outputs with
a deliberately diverse four-choice menu at every Viterbi decision.

\begin{table}[t]
\centering
\caption{Additional exhaustive properties of the XOR-Trellis generator.}
\label{tab:XOR-Trellis-properties}
\begin{tabular}{lc}
\toprule
Property & Value\\
\midrule
Count per payload code & 4096\\
Mean & 0\\
Standard deviation & 2.9262\\
Mean absolute value & 2.2500\\
$P(q=0)$ & 12.5\%\\
$P(|q|\geq 4)$ & 25.0\%\\
$P(|q|=6)$ & 12.5\%\\
Distinct numeric local menus & 4\\
Menus with duplicate numeric values & 0\%\\
Menus missing either sign & 0\\
Minimum spacing within every menu & 1.5\\
State bits with nonzero numeric influence & 16 of 16\\
\bottomrule
\end{tabular}
\end{table}

\begin{figure}[t]
    \centering
    \includegraphics[width=\columnwidth]{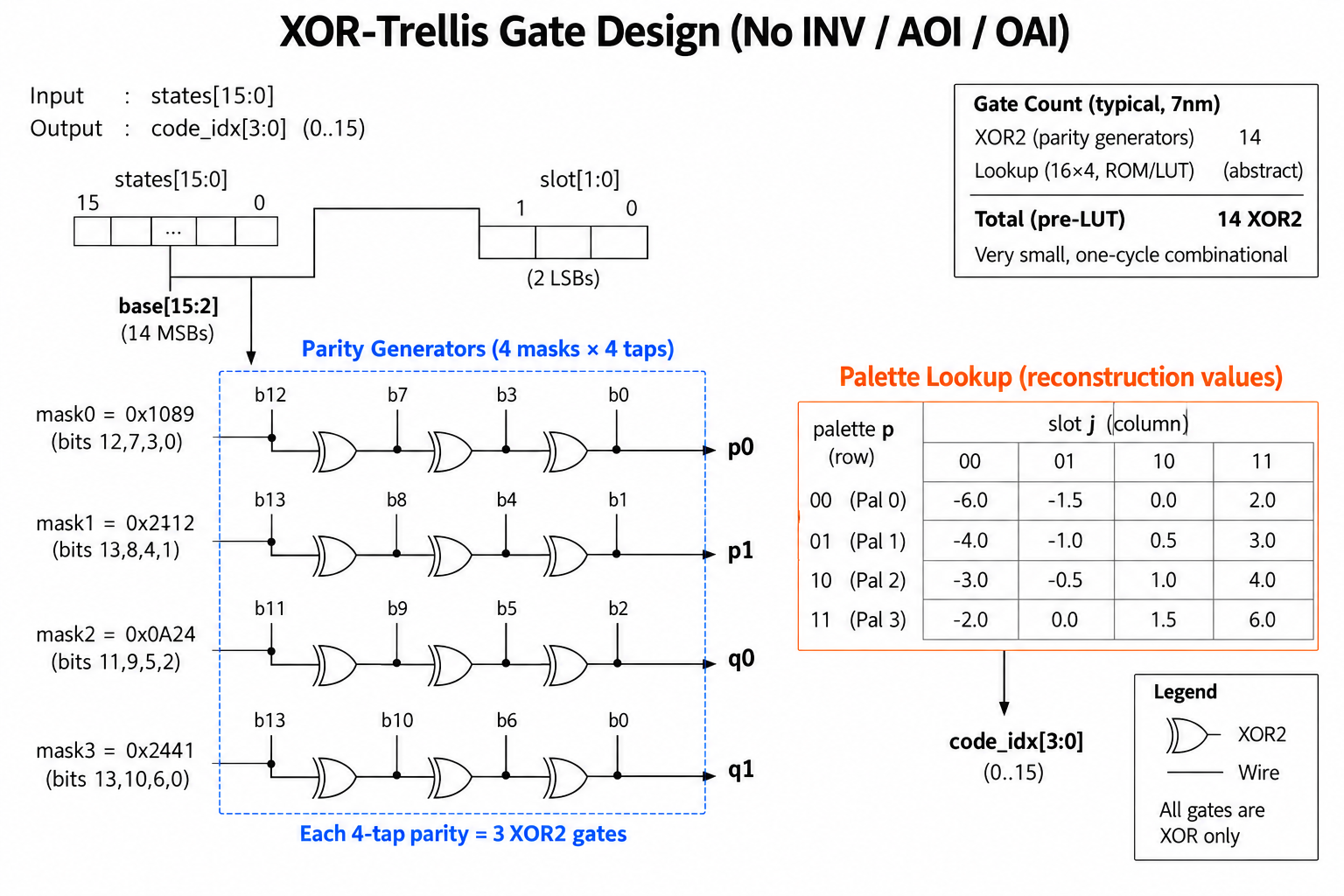}
    \caption{
    Gate-level implementation of the XOR-Trellis generator.
    Four masked parity functions are computed from the 14-bit history using XOR2 gates, producing the two-bit palette selector and two permutation bits.
    Together with the two transition-slot bits, these signals index a fixed $4\times4$ palette lookup to produce one of the 16 FP4 payload codes.
    The parity logic requires 14 XOR2 gates before the lookup, resulting in a compact combinational implementation.
    }
    \label{fig:xor_trellis_gate_design}
\end{figure}

\paragraph{Gate-level implementation.}
Figure~\ref{fig:xor_trellis_gate_design} shows that the XOR-Trellis mapping can be implemented with lightweight combinational logic. The four masked parity functions depend only on the shared 14-bit history and require three XOR2 gates per four-tap parity. The resulting palette and permutation bits, together with the two transition-slot bits, address a fixed $4\times4$ lookup table containing the 16 FP4 payload codes. Thus, the mapping introduces only a small amount of XOR logic in addition to the fixed lookup and does not require multipliers, rotations, or general arithmetic.
\paragraph{Offline optimization and unchanged online representation.}
The curvature-aware metric is used only during offline quantization. The diagonal factors $D$ are produced by the LDL decomposition and are used solely to weight candidate branch costs during Viterbi search. They are neither stored with the deployed model nor used by the inference-time decoder. Consequently, the compressed representation and the XOR-Trellis dequantization logic remain unchanged.

The two components therefore operate at different stages of the pipeline:
\begin{equation*}
\underbrace{\text{curvature-aware trellis search}}_{\text{offline quantization}}
\;\longrightarrow\;
\underbrace{\text{2-bit trellis sequence}}_{\text{stored model}}
\;\longrightarrow\;
\underbrace{\text{XOR-Trellis FP4 dequantizer}}_{\text{online inference}}.
\end{equation*}
This separation enables model-aware offline optimization without increasing the complexity of the runtime decoder.
\end{document}